%% file: main.tex
\documentclass[10pt,twocolumn,letterpaper]{article}

\usepackage{iccv}              

\input{preamble}

\definecolor{iccvblue}{rgb}{0.21,0.49,0.74}
\usepackage[pagebackref,breaklinks,colorlinks,allcolors=iccvblue]{hyperref}
\usepackage{xcolor}
\usepackage{graphicx}
\usepackage{adjustbox}
\usepackage{siunitx}
\usepackage{tcolorbox}
\usepackage[accsupp]{axessibility}
\usepackage{multirow}
\tcbuselibrary{skins, breakable}

\definecolor{gcyan}{RGB}{47, 229, 192}
\newcommand{\D}[1]{\textcolor{gcyan}{#1}}
\newcommand{\T}[1]{\textcolor{red}{#1}}

\usepackage{lineno}
\newcommand{\SV}[1]{}
\newcommand{\YSR}[1]{}
\newcommand{\AAD}[1]{}

\def\paperID{5} 
\def\confName{ICCV}
\def\confYear{2025}

\title{
  \adjustbox{valign=c}{\includegraphics[height=28pt]{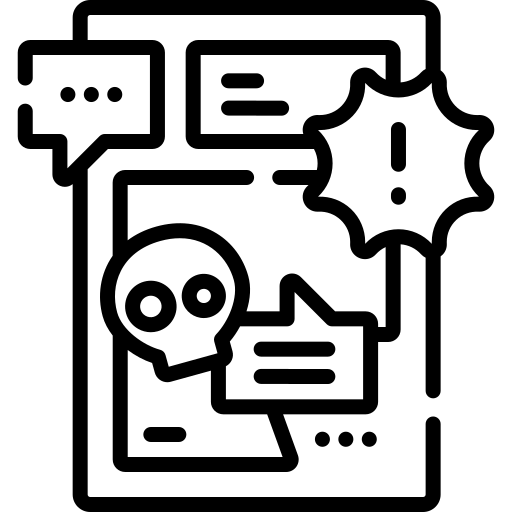}}~%
  Re:Verse - Can Your VLM Read a Manga?
}

\author{Aaditya Baranwal\textsuperscript{$\dagger$}\\
\small University of Central Florida\\
\small Orlando, FL, USA\\
{\tt\small aaditya.baranwal@ucf.edu}
\and
Madhav Kataria\\
\small Indian Institute of Technology, Jodhpur\\
\small Karwar, RJ, IN\\
{\tt\small b23ch1025@iitj.ac.in}
\and
Naitik Agarwal\\
\small Indian Institute of Technology, Varanasi\\
\small Varanasi, UP, IN\\
{\tt\small naitikagrawal838@gmail.com}
\and
Yogesh Singh Rawat\\
\small University of Central Florida\\
\small Orlando, FL, USA\\
{\tt\small yogesh@crcv.ucf.edu}
\and
Shruti Vyas\\
\small University of Central Florida\\
\small Orlando, FL, USA\\
{\tt\small shruti@ucf.edu}
}

\begin{document}

\twocolumn[{
\maketitle
\vspace{-1em}
\begin{center}
    \includegraphics[width=0.96\textwidth]{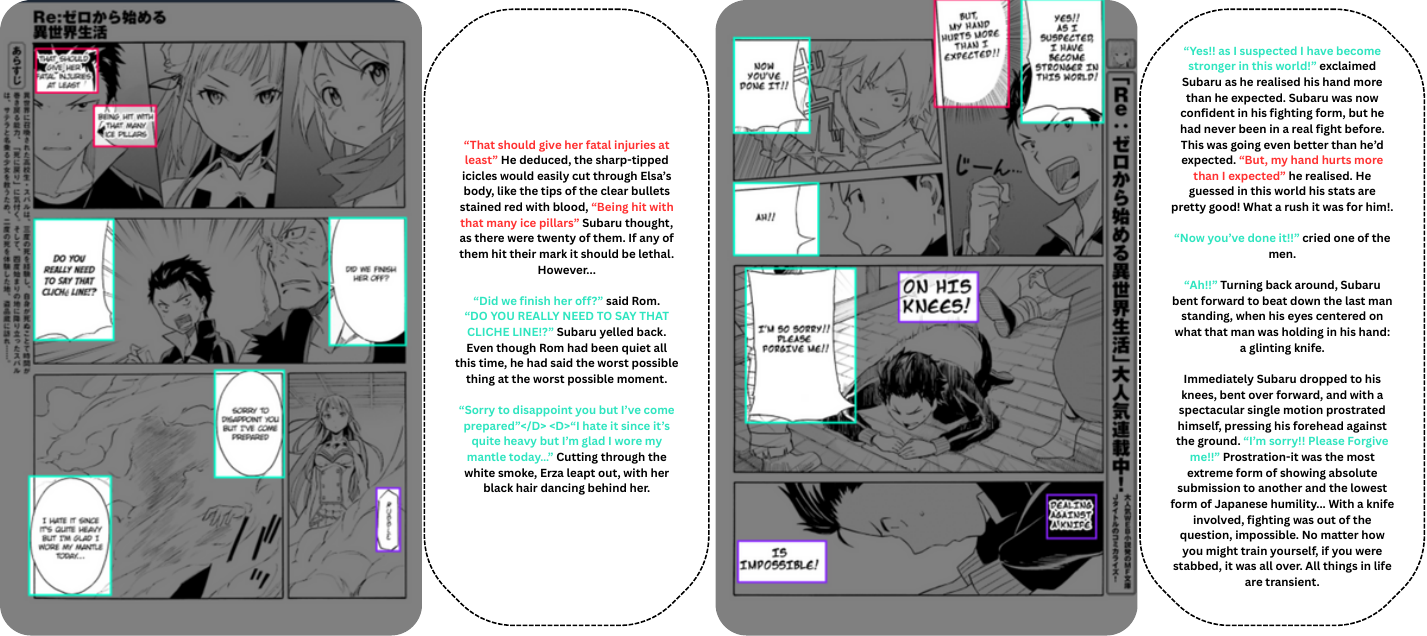}
    \captionof{figure}{\textbf{Re:Verse Multimodal Annotation Examples.} 
    Two representative examples from our dataset demonstrate the fine-grained alignment between visual manga content and narrative text. Each example shows the original manga page (left) paired with its corresponding aligned narrative text (right), where \D{\texttt{<D></D>}} tags indicate spoken dialogue (displayed in cyan) and \T{\texttt{<T></T>}} tags indicate internal thoughts (displayed in red). This precise cross-modal correspondence enables systematic evaluation of vision-language models' ability to understand sequential narrative structure, character consistency, and temporal progression. The semantic markup facilitates a comprehensive assessment of both surface-level text recognition and deep narrative comprehension capabilities across extended manga sequences.
    }    
    \label{fig:narrative_visual_grounding}
\end{center}
\vspace{1em}
}]

\maketitle
\input{sec/0_abstract}    
\input{sec/1_intro}

\input{sec/2_relatedwork}

\input{sec/3_datex}
\input{sec/4_analysis}
\input{sec/5_conclusion}
\input{sec/6_limitations}
{   \clearpage
    \small
    \bibliographystyle{ieeenat_fullname}
    \bibliography{main}
}

\input{sec/X_suppl}

\end{document}

%% file: preamble.tex
\usepackage[T1]{fontenc}
\usepackage[utf8]{inputenc}

\usepackage{subcaption}      
\usepackage{booktabs}        
\usepackage{array}           
\usepackage{multirow}        
\usepackage{amsmath}         
\usepackage{amssymb}         
\usepackage{textcomp}        

\DeclareRobustCommand{\up}{\ensuremath{\uparrow}}

%% file: sec/0_abstract.tex
\begin{abstract}
Current Vision Language Models (VLMs) demonstrate a critical gap between surface-level recognition and deep narrative reasoning when processing sequential visual storytelling. Through a comprehensive investigation of manga narrative understanding, we reveal that while recent large multimodal models excel at individual panel interpretation, they systematically fail at temporal causality and cross-panel cohesion, core requirements for coherent story comprehension. We introduce a novel evaluation framework that combines fine-grained multimodal annotation, cross-modal embedding analysis, and retrieval-augmented assessment to systematically characterize these limitations. 

Our methodology includes (i) a rigorous annotation protocol linking visual elements to narrative structure through aligned light novel text, (ii) comprehensive evaluation across multiple reasoning paradigms, including direct inference and retrieval-augmented generation, and (iii) cross-modal similarity analysis revealing fundamental misalignments in current VLMs' joint representations. Applying this framework to Re:Zero manga across 11 chapters with 308 annotated panels, we conduct the first systematic study of long-form narrative understanding in VLMs through three core evaluation axes: generative storytelling, contextual dialogue grounding, and temporal reasoning. Our findings demonstrate that current models lack genuine story-level intelligence, struggling particularly with non-linear narratives, character consistency, and causal inference across extended sequences. This work establishes both the foundation and practical methodology for evaluating narrative intelligence, while providing actionable insights into the capability of deep sequential understanding of Discrete Visual Narratives beyond basic recognition in Multimodal Models.

Project Page: \href{https://re-verse.vercel.app}{https://re-verse.vercel.app}

Github: \href{https://github.com/eternal-f1ame/Re-Verse}{https://github.com/eternal-f1ame/Re-Verse}
\end{abstract}

%% file: sec/1_intro.tex
\vspace{-2em}
\section{Introduction}
\label{sec:intro}
Manga represents a rich and distinctive form of multimodal storytelling that blends expressive artwork, diverse panel layouts, and text embedded directly within images. Unlike standard natural images, understanding manga requires following visual and textual cues across multiple panels, interpreting characters' emotions, and tracking complex storylines~\citep{cohn2013visual}. These narratives often include non-linear timelines, inner thoughts, and symbolic visuals, making them uniquely challenging for computational models~\citep{yefymenko2022multimodality}.

As vision-language models (VLMs) continue to advance multimodal understanding~\citep{radford2021learning,li2022blip,liu2024visual}, applying them to manga opens exciting possibilities for supporting human creativity. Models that can understand manga could help creators edit, summarize, or expand their stories, acting as intelligent assistants that comprehend narrative flow~\citep{shen2023maru}. To be truly useful, such models must follow context over many panels, recognize characters consistently, and reason about subtle visual and textual elements in a human-like manner. However, most existing benchmarks for manga understanding are limited in scope, focusing primarily on multiple-choice or classification-based tasks that operate at individual panel levels~\citep{Aizawa_2020,ikuta2024mangaubmangaunderstandingbenchmark,baek2025mangavqamangalmmbenchmarkspecialized}, without capturing broader narrative understanding or the full complexity of manga storytelling. This complexity often relies on events spread across multiple pages, emotional shifts, and implied information that requires sophisticated reasoning across extended sequences~\citep{iyyer2017amazing,cohn2012peanuts}. To support the next generation of intelligent systems for manga, there is a critical need for benchmarks that evaluate models on their ability to follow long-range context, generate coherent summaries, and make narrative predictions for this intermediate modality of discrete visual narratives, sequences or exerpts.~\citep{vivoli2024comixcomprehensivebenchmarkmultitask}.

\begin{figure}[t]
    \centering
    \includegraphics[width=\linewidth]{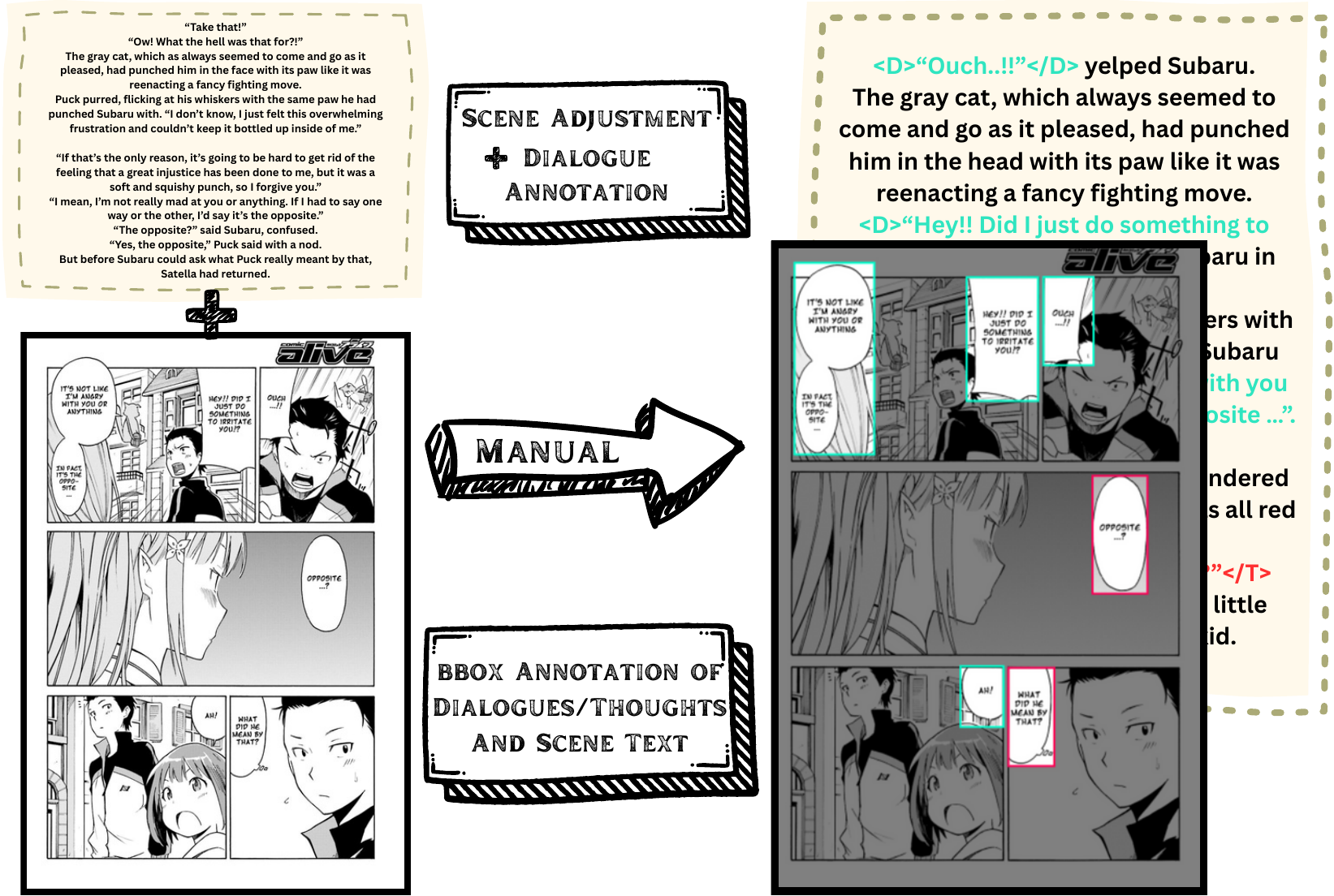}
    \caption{
    \textbf{Re:Verse curation and annotation pipeline.} Raw manga panels and light novel text are aligned via manual bounding box annotation, text classification (\texttt{<D>}, \texttt{<T>}), and semantic grounding, yielding synchronized visual, spatial, and narrative structure.
    }
    \label{fig:performance}
\end{figure}

To address this fundamental gap, we introduce Re:Verse, a comprehensive benchmark for sequential narrative understanding in manga, designed to evaluate whether models can track and interpret story progression across entire chapters. Re:Verse is constructed from Arc 1 of Re:Zero~\citep{rezero_ln_vol1, rezero_manga_vol1}, a popular manga known for its intricate storytelling involving reincarnation loops and temporal resets. These narrative elements make Re:Zero an ideal testbed for evaluating long-range reasoning and panel-to-panel coherence in non-linear narratives~\citep{shangguan2024tomato}. Our benchmark contains 308 manually annotated panels from 11 chapters, with each panel including: (i) fine-grained bounding boxes and semantic tags for dialogue, thoughts, and key scene elements, and (ii) manually curated alignment with corresponding Re:Zero light novel passages, providing precise grounding (for detailed schematic procedure refer Figure~\ref{fig:performance}). 

We evaluate both open-source and proprietary models (with major focus on opensourced VLMs)~\citep{zhu2023minigpt,bai2023qwen,team2023gemini,openai2024gpt4ocard} across a comprehensive set of narrative tasks. Our findings reveal that while current models can extract surface-level semantics, they systematically struggle with maintaining narrative coherence, understanding character motivations, and handling temporal dependencies—particularly in stories with non-linear structures like Re:Zero~\citep{cater2021fabula,wright2013evaluating}.

\noindent The contributions of this paper are:
\begin{itemize}
    \item We introduce \textbf{Re:Verse}, a novel benchmark for manga narrative understanding, featuring 308 annotated panels with aligned visual and textual data that enables systematic evaluation of long-form narrative comprehension.
    \item We conduct comprehensive evaluation of state-of-the-art vision-language models across three challenging tasks: story generation, dialogue grounding, and temporal reasoning, revealing current fundamental limitations.
    \item We provide detailed analysis of model performance that characterizes key gaps in long-range coherence and multimodal narrative comprehension, establishing a foundation for future research in sequential visual storytelling.
\end{itemize}

%% file: sec/2_relatedwork.tex
\section{Related Work}
\label{sec:relwork}

\begin{table*}[t]
\centering
\footnotesize
\caption{
Comparison of \textbf{Re:Verse} with existing benchmarks. Re:Verse provides chapter-length narrative evaluation with aligned textual grounding, enabling systematic assessment of story comprehension. 
\textit{Task abbreviations:} NG = Narrative Grounding, DA = Dialogue Attribution, TR = Temporal Reasoning. 
\textit{Evaluation types:} HE = Human Evaluation, AE = Automatic Evaluation, LLM-Based Evaluation.
}
\resizebox{\textwidth}{!}{%
\begin{tabular}{l|cccccc}
\toprule
\textbf{Benchmark} & \textbf{Temporal Span} & \textbf{Tasks} & \textbf{Evaluation Focus} & \textbf{Aligned Text} & \textbf{Evaluation Methods} \\
\midrule
Manga109~\citep{Aizawa_2020} & Single & SP, OCR & Structural Analysis & -- & AE \\
MangaUB~\citep{ikuta2024mangaubmangaunderstandingbenchmark} & Short-seq & VQA, PC & Panel Relations & -- & AE (CircularEval) \\
MangaVQA~\citep{baek2025mangavqamangalmmbenchmarkspecialized} & 2-Page & VQA & Spread Understanding & -- & AE \\
CoMix~\citep{vivoli2024comixcomprehensivebenchmarkmultitask} & Single & Multi-task & Integrated Reasoning & -- & AE \\
\midrule
\textbf{Re:Verse (Ours)} & \textbf{Chapter-length} & \textbf{NG, DA, TR} & \textbf{Narrative Intelligence} & \textbf{\checkmark} & \textbf{HE, AE, LE} \\
\bottomrule
\end{tabular}
}
\label{tab:benchmark_comparison}
\end{table*}

Computational analysis of comics and manga has evolved from structural detection to narrative comprehension, progressing through three stages: foundational datasets, task-specific benchmarks, and modern narrative understanding with Large Multimodal Models (LMMs).

\noindent\textbf{Foundational Datasets for Structural Analysis}

Early research established large-scale datasets for fundamental tasks. Manga109 remains influential, providing 109 volumes with detailed annotations for panels, text, characters, and faces~\citep{Aizawa_2020}, driving manga-specific structural analysis and computational comic understanding~\citep{cohn2013visual}. To diversify artistic styles, datasets like COMICS, eBDtheque~\citep{eBDtheque2013}, and DCM772~\citep{vivoli2024comicsdatasetsframeworkmix} introduced American and European comics, though often with automated text extraction quality issues. COMICS Text+ addressed this with high-quality OCR benchmarks for Western comics~\citep{soykan2022comprehensivegoldstandardbenchmark}. These efforts established foundations for parsing visual grammar and understanding multimodal sequential storytelling~\citep{yefymenko2022multimodality}.

\noindent\textbf{Task-Specific Benchmarks and Integrated Reasoning}

Subsequent work developed specialized benchmarks for integrated tasks. The Comic Onomatopoeia (COO) dataset extended Manga109 for recognizing stylized sound effects~\citep{baek2022coocomiconomatopoeiadataset}, while foundational work on "closure" between panels~\citep{iyyer2017amazing} established theoretical frameworks for sequential image comprehension. CoMix introduced comprehensive multi-task evaluation including speaker identification and dialogue generation~\citep{vivoli2024comixcomprehensivebenchmarkmultitask}, notably incorporating American comics to reduce manga over-representation and shifting focus from single-task to multi-task reasoning. Complementary efforts include M2C for manga complement generation~\citep{guo2023m2c} and MaRU for manga understanding~\citep{shen2023marumangaretrievalunderstanding}.

\begin{figure*}[t]
    \centering
    \begin{subfigure}[c]{0.49\linewidth}
        \centering
        \includegraphics[width=\linewidth]{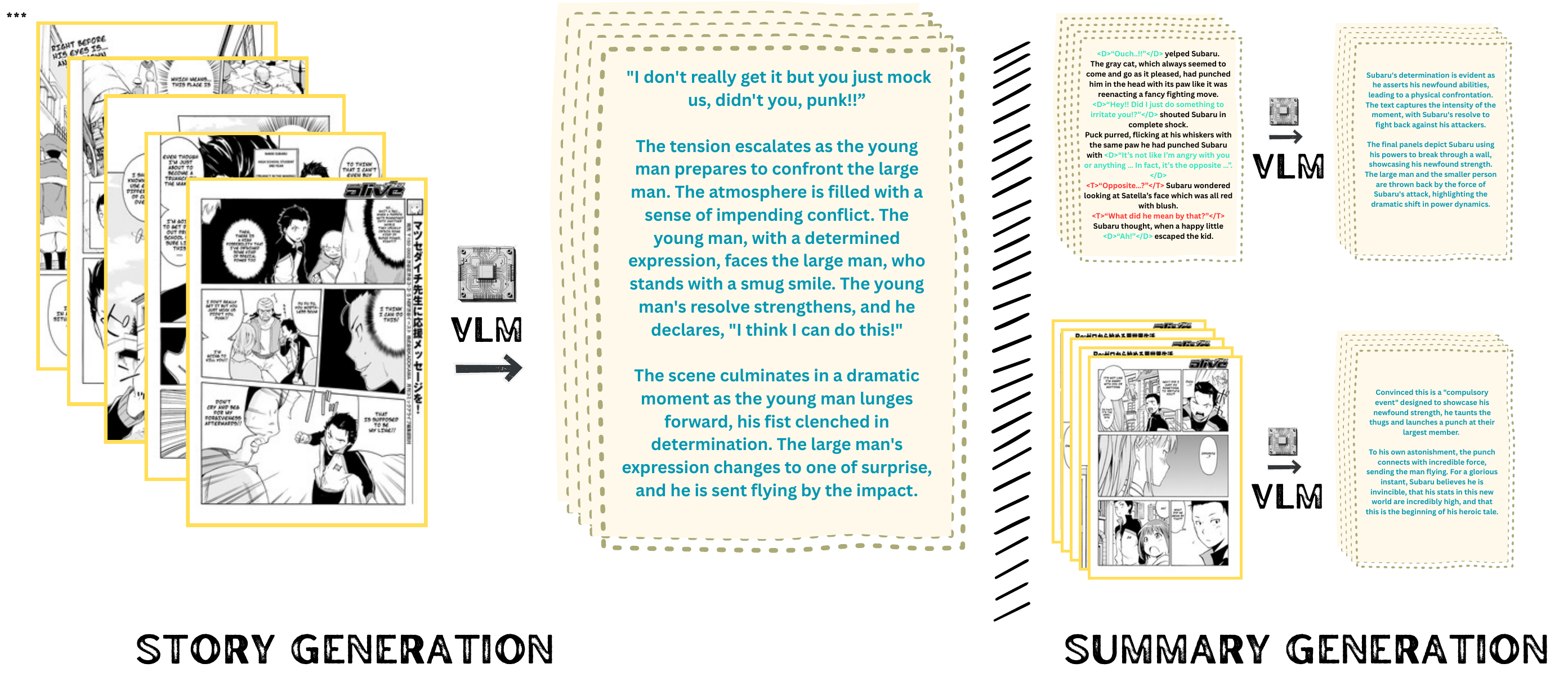}
        \caption{Generative Setup: Story and Summary Generation.}
        \label{fig:experiment_1}
    \end{subfigure}
    \hfill
    \begin{subfigure}[c]{0.49\textwidth}
        \centering
        \includegraphics[width=\linewidth]{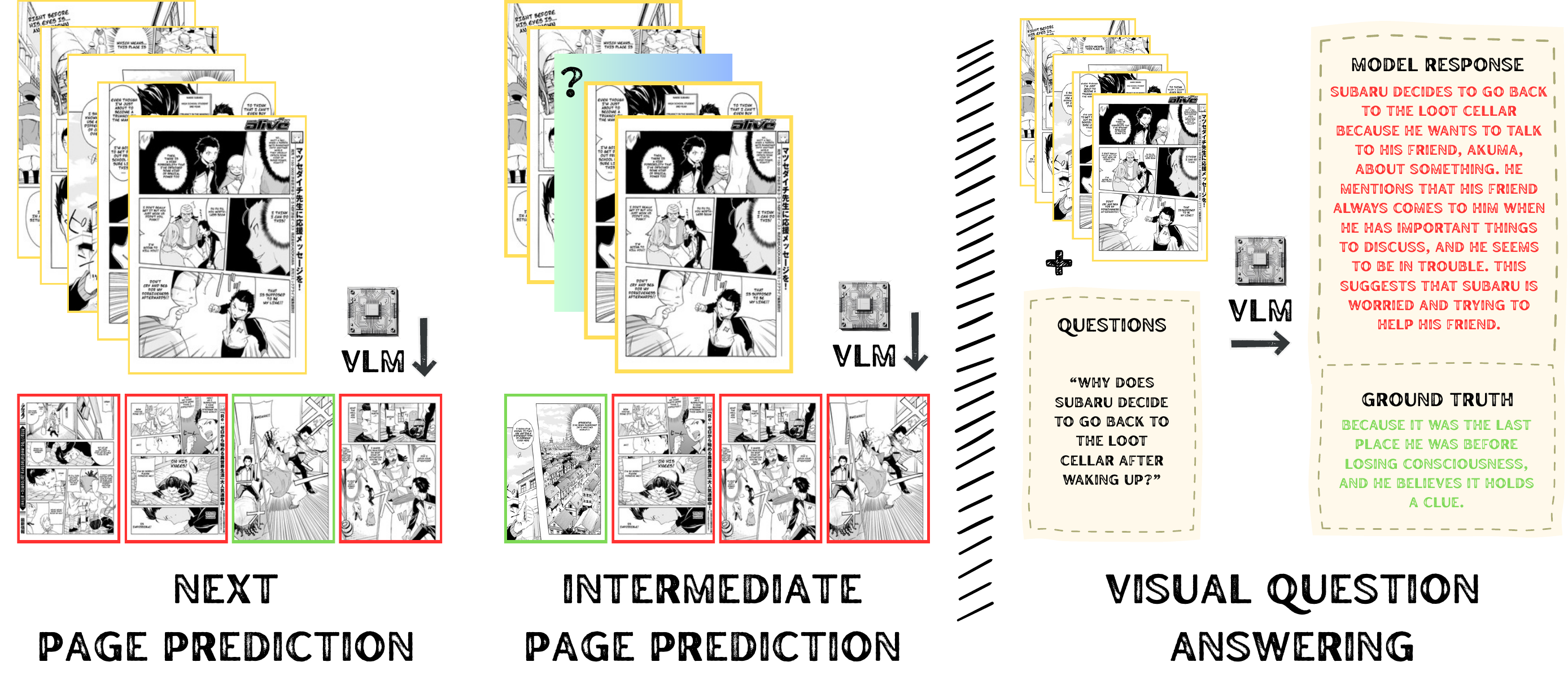}
        \caption{Temporal Reasoning: Page-Predictions and Visual Question-Answering.}
        \label{fig:experiment_2}
    \end{subfigure}
    \caption{
        \textbf{Overview of experimental setups used to evaluate narrative understanding.} 
        \subref{fig:experiment_1} depicts the \textit{Generative Tasks}, where a Vision-Language Model (VLM) is prompted to either (i) synthesize detailed narratives (\textbf{Story Generation}) or (ii) produce concise plot summaries (\textbf{Summary Generation}) 
        \subref{fig:experiment_2} shows the \textit{Temporal Reasoning Tasks}, (i) \textbf{Next Page Prediction}, where VLM selects the correct continuation of a story, (ii) \textbf{Intermediate Page Prediction} tests inference over missing narrative, and (iii) \textbf{VQA} for overall narrative understanding.
    }
    \label{fig:experiments}
\end{figure*}

\begin{figure}
    \centering
    \includegraphics[width=\linewidth]{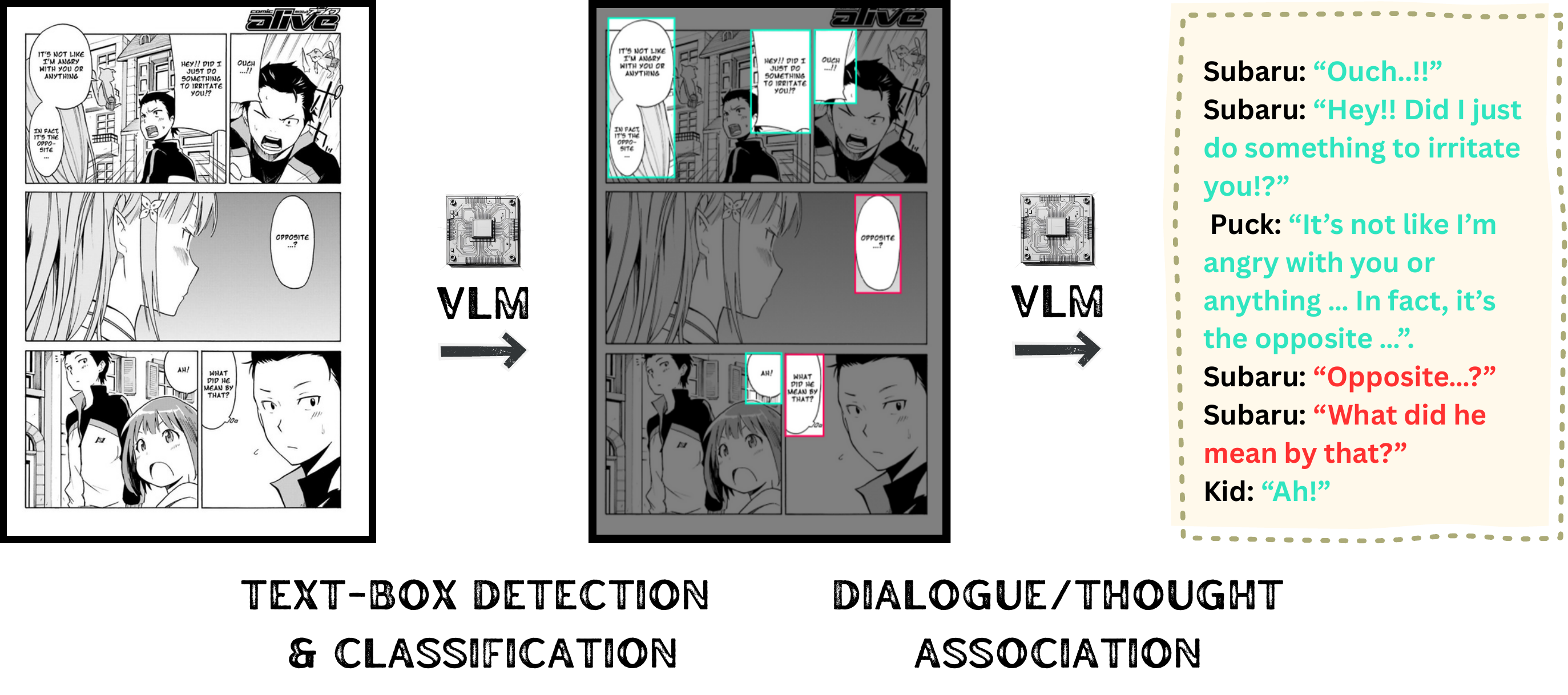}
    \caption{\textbf{Grounding Setup:} Text Detection, Classification, and Association with characters to evaluate in-page understanding.}
    \label{fig:experiment_3}
\end{figure}

\noindent\textbf{Towards Narrative Comprehension with LMMs}

The advent of LMMs~\citep{radford2021learning,li2022blip,liu2024visual,zhu2023minigpt,bai2023qwen} catalyzed deeper narrative understanding evaluation. MangaUB performed fine-grained LMM analysis on single-panel recognition and multi-panel comprehension, identifying panel connections as key challenges~\citep{ikuta2024mangaubmangaunderstandingbenchmark}. MangaVQA advanced this with VQA tasks on two-page spreads, better emulating human reading and requiring cross-context synthesis~\citep{baek2025mangavqamangalmmbenchmarkspecialized}. Recent work explored sequential image-to-text reasoning~\citep{villegas2025imagechain} and cross-modal alignment~\citep{kim2022transferring,li2024vision}.

"From Panels to Prose" represents the most ambitious goal, generating full literary prose from manga pages including dialogue, descriptions, and actions~\citep{sachdeva2025panelsprosegeneratingliterary}. Parallel developments in story evaluation~\citep{cater2021fabula,wright2013evaluating,yi2025scorestorycoherenceretrieval} and narrative coherence assessment~\citep{chun2024aistorysimilarity,yang2024makesgoodstorymeasure,openmeva2021,chhun2022humancriteriaautomaticmetrics} provided theoretical frameworks, while temporal reasoning benchmarks like TOMATO~\citep{shangguan2024tomato} addressed temporal causality challenges. The field benefits from cognitive science research on visual narrative structure~\citep{cohn2012peanuts} and comprehensive surveys of multimodal capabilities~\citep{li2024survey,zhang2023vision,lee2024vhelm}.

While these benchmarks have pushed the field from structural detection towards contextual reasoning, they predominantly operate on short sequences or isolated tasks (Table~\ref{tab:benchmark_comparison}). A significant gap remains in evaluating the ability of LMMs to maintain **long-form narrative coherence** across entire chapters or story arcs. Our work, Re:Verse, is designed to fill this specific void. By providing a continuous, chapter-length dataset with aligned narrative text, it introduces the first benchmark focused on assessing temporal causality, character consistency, and story-level cohesion—aspects not systematically measured by prior work.

%% file: sec/3_datex.tex
\section{Re:Verse Benchmark}
\label{sec:benchmark}

To address the fundamental limitations in existing benchmarks for sequential visual narratives, we introduce \textbf{Re:Verse}, a comprehensive benchmark designed to evaluate VLMs' ability to understand long-form manga narratives. Unlike existing benchmarks that focus on isolated panels or short sequences, Re:Verse systematically evaluates three critical aspects of narrative intelligence: story synthesis, character grounding, and temporal reasoning.

\subsection{Dataset Construction}

We construct our benchmark from Volume 1 of \textit{Re:Zero - Starting Life in Another World}, chosen for its complex narrative structure featuring time-loop mechanics, diverse character interactions, and sophisticated visual storytelling. The dataset comprises 308 meticulously annotated manga pages spanning 11 chapters, with each page containing:

\textbf{Spatial Annotations.} Every text bubble, thought bubble, and scene element is precisely localized with bounding boxes and semantic tags. This fine-grained spatial grounding enables evaluation of visual-textual integration.

\textbf{Narrative Alignment.} An expert annotator with comprehensive series knowledge manually adapted corresponding light novel passages to achieve one-to-one narrative correspondence with manga content. This alignment provides ground truth for story-level evaluation tasks.

\textbf{Sequential Structure.} Pages are organized to preserve narrative flow and temporal dependencies, enabling systematic evaluation of long-range coherence and causal reasoning across extended sequences as in the original work.

\subsection{Analysis Framework}

Our analysis framework systematically probes narrative understanding through three task categories (Figure~\ref{fig:experiments}):

\textbf{Story Synthesis Tasks.} We evaluate models' holistic narrative understanding through two generative tasks. \textit{Story Generation} requires models to transform entire manga chapters into coherent prose narratives. \textit{Cross-modal Summarization} evaluates models on summarizing both manga pages and their corresponding light novel text independently, quantifying comprehension and inference gaps and misses introduced in the visual storytelling modality.

\textbf{Character Grounding Tasks.} \textit{Dialogue Extraction and Attribution} task requires models to spatially locate text elements, classify their semantic types (dialogue vs. thoughts), and correctly attribute them to characters using visual cues.

\textbf{Temporal Reasoning Tasks.} We frame temporal understanding through objective measures. \textit{Next Page Prediction $(NPP)$} and \textit{Intermediate Page Prediction $(IPP)$} evaluates models' ability to infer missing narrative events by predicting pages that bridge non-consecutive sequences. We predict the 6th and 11th page given the 5 and 10 consecutive prior pages, respectively, for $NPP$, and predict the missing page in a 2-missing-2 page and 3-missing-3 page setup for $IPP$. \textit{Visual Question Answering} employs 55 complex questions requiring information across multiple pages.

\subsection{Evaluation Methodology}

Our evaluation protocol builds upon and extends prior frameworks such as SCORE~\cite{yi2025scorestorycoherenceretrieval} and AIStorySimilarity~\cite{chun2024aistorysimilarity}. Unlike binary or pass/fail approaches, our methodology introduces proportional penalty systems, minimum score thresholds, and state-based incremental evaluation, enabling fine-grained differentiation of narrative quality. We employ both an LLM (GPT-4o \citep{openai2024gpt4ocard}) and automated metrics (BERTScore, ROUGE, STTR, lexical density, NER density), with statistical significance testing and embedding-based cross-modal alignment (details in supplementary). Temporal reasoning tasks use accuracy measures within VQA frameworks. Cross-modal alignment is assessed through correlation analysis and embedding quality metrics using four vision encoders (BLIP, CLIP, SIGLIP, ALIGN).

\textbf{Data Processing Pipeline.} All manga pages undergo standardized preprocessing: resized to $224\times width$ pixels while maintaining aspect ratio. Text annotations preserve dialogue tags $(⟨D⟩⟨/D⟩)$ and thought tags $(⟨T⟩⟨/T⟩)$ to maintain narrative structure distinctions crucial for evaluation. Character and entity names undergo manual validation using spaCy's named entity recognition enhanced with Re:Zero story-specific terminology (pronouns, salutations, honorifics, etc.).

\textbf{Metric Standardization.} NER density calculation uses character mention ratios to total word count. STTR employs sliding window approach (window size 50, step size 10) following standard text quality assessment. BERTScore utilizes bert-base-uncased with default settings for robust semantic similarity assessment. All results include confidence intervals ensuring statistical reliability.

This comprehensive evaluation framework enables systematic characterization of current VLM limitations in sequential visual narrative understanding, providing actionable insights for developing models with genuine narrative intelligence (understanding and generation) capabilities.

%% file: sec/4_analysis.tex
\section{Analysis}
\label{sec:analysis}

We analyze VLM performance across story synthesis, visual-textual integration, and temporal reasoning. Our findings reveal systematic challenges in processing manga's \textit{discrete temporal structure}, where narrative meaning emerges from discontinuous visual sequences. This analysis identifies the \textit{inferent gap}---narrative discontinuity between pages requiring inference of missing story elements---as a fundamental bottleneck for current VLMs.

\textbf{Model Selection Rationale.} We focus on open-source VLMs rather than proprietary models for several critical reasons: (1) \textit{Data contamination prevention}---Re:Zero's popularity likely exposes proprietary models to extensive training data contamination, making unbiased evaluation impossible; (2) \textit{Reproducibility requirements}---academic benchmarks require stable, accessible models that enable reproducible research; (3) \textit{Evaluation scale and cost}---comprehensive assessment across 308 pages with multiple tasks would be prohibitively expensive with proprietary APIs; and (4) \textit{Democratization of comics understanding}---as open-source VLMs excel in video understanding, it is essential to democratize sequential visual narrative capabilities. This focus enables actionable insights for the broader research community while ensuring evaluation integrity.

\subsection{Story Synthesis: Consistency Failures}

Our story generation evaluation reveals systematic narrative coherence limitations across all models. Table~\ref{tab:story-generation} summarizes performance across six metrics, showing consistent gaps compared to human-authored narratives.

\begin{table}
\caption{Story generation performance across VLMs showing systematic character consistency failures and quality degradation compared to human narratives. NER density measures character mention frequency; STTR measures narrative tone and style.}
\resizebox{\linewidth}{!}{%
\footnotesize
\begin{tabular}{|lcccccc|}
\toprule
\textbf{Model} & \textbf{NER Density} & \textbf{STTR} & \textbf{Length} & \textbf{F1} & \textbf{ROUGE-1} & \textbf{Lexical Density} \\
\midrule
InternVL3-2B & 0.023 & 0.71 & 89.4 & 0.803 & 0.143 & 0.54 \\
Ovis2-2B & 0.012 & 0.72 & 102.7 & 0.765 & 0.135 & 0.55 \\
Qwen2.5-VL-3B & 0.027 & 0.78 & 96.8 & 0.814 & 0.152 & 0.55 \\
InternVL3-8B & 0.012 & 0.81 & 95.1 & 0.817 & 0.151 & 0.56 \\
Ovis2-8B & 0.010 & 0.78 & 98.3 & 0.805 & 0.149 & 0.56 \\
Qwen2.5-VL-7B & 0.015 & 0.80 & 101.2 & 0.823 & 0.156 & 0.55 \\
InternVL3-14B & 0.009 & 0.81 & 97.6 & 0.819 & 0.165 & 0.57 \\
Ovis2-16B & 0.019 & 0.79 & 93.4 & 0.831 & 0.173 & 0.56 \\
\midrule
\textbf{Gold Standard} & 0.087 & 0.82 & 156.2 & 1.000 & 1.000 & 0.61 \\
\bottomrule
\end{tabular}
}
\vspace{-0.4em}
\label{tab:story-generation}
\end{table}

The most critical finding is systematic character consistency failure. Named Entity Recognition (NER) density ranges from 0.009 to 0.027 across VLMs, representing a 3-10x gap compared to the gold standard of 0.087. Even the best performing model (Qwen2.5-VL-3B) falls 69\% short of human performance, indicating models cannot maintain character references across extended sequences. Quality metrics reveal additional degradation: Surface-form Type-Token Ratio (STTR) values range from 0.71 to 0.81 compared to 0.82 for human narratives, while generated content is systematically shorter (89-102 vs. 156 words) with lower lexical density. These patterns indicate models generate repetitive, formulaic content rather than nuanced narratives capturing rich character interactions.

\textbf{Cross-modal Processing.} Our cross-modal analysis (Table. \ref{tab:comprehensive-performance} Summarization) compares corresponding content processing through text versus manga pages, revealing consistently close scores for automated metrics with gaps of just ranging from 1.1-3.2\% BERTScore F1 points. Evidently, the close vicinity of the two scores establishes that automated metrics alone are nonsensical in the evaluation of narrative summarisation from visual content.

\subsection{Character Grounding: Attribution Failures}

We conducted a two-stage evaluation of dialogue extraction and character affiliation, revealing a critical breakdown in character-dialogue relationships. Table~\ref{tab:character_attribution} shows volume-wide performance across all models, aggregated from chapter-level evaluations.

\begin{table}
\caption{Character dialogue extraction and attribution performance demonstrating the critical disconnect between spatial text detection and character-dialogue grounding across VLMs.}
\resizebox{\linewidth}{!}{%
\footnotesize
\begin{tabular}{|lcccc|}
	\toprule
	\textbf{Model} & \textbf{Recall} & \textbf{Precision} & \textbf{F1-Score} & \textbf{Character Accuracy} \\
\midrule
Qwen2.5-VL-3B & 23.75\% & 21.84\% & 22.25\% & 1.11\% \\
Ovis-8B & 17.47\% & 9.89\% & 9.84\% & 0.85\% \\
InternVL-14B & 29.17\% & 20.65\% & 23.08\% & 0.00\% \\
\bottomrule
\end{tabular}
}
\label{tab:character_attribution}
\end{table}

The results reveal a catastrophic disconnect between dialogue extraction and character understanding. While models achieve modest F1-scores for dialogue extraction (9.84\%-23.08\%), character attribution accuracy is near-zero (0.00\%-1.11\%). Most strikingly, InternVL-14B completely fails to correctly attribute any dialogue to characters despite achieving the highest extraction F1-score of 23.08\%, while Qwen2.5-VL-3B manages only 1.11\% character accuracy.

These character attribution failures provide direct mechanistic evidence for the systematic character consistency gaps observed in our story generation experiments. The near-zero character accuracy scores demonstrate that current VLMs cannot establish the integrated visual-textual understanding necessary to ground dialogue in character identities---a fundamental requirement.

\subsection{Visual-Textual Integration}

We evaluate VLM performance on visual-textual integration and temporal reasoning tasks. Manga represents a unique \textit{middle modality} where visual and textual information are spatially integrated, requiring simultaneous processing across multiple cognitive dimensions (Table~\ref{tab:comprehensive-performance}).

\begin{table*}[t]
\centering
\caption{\textbf{Comprehensive Performance across Visual-Textual Integration and Temporal Reasoning tasks.} Cross-modal penalties reveal processing degradation (TI = Text Input, VI = Visual Input), spatial localization succeeds but text extraction fails, while InternVL3-8B uniquely benefits from longer context and larger context gaps improve intermediate prediction.}
\label{tab:comprehensive-performance}

\setlength{\tabcolsep}{4pt}
\sisetup{
  mode=text,
  table-align-text-post=false,
  text-family-to-math=true
}

\footnotesize 
\begin{tabular}{
  |l|
  S[table-format=1.3]
  S[table-format=1.3]|
  S[table-format=1.3]
  S[table-format=1.3]
  S[table-format=1.3]
  S[table-format=1.3]|
  S[table-format=2.1]
  S[table-format=2.1]
  S[table-format=1.3]
  S[table-format=1.3]|
  S[table-format=2.1]
  S[table-format=2.1]
  S[table-format=1.3]
  S[table-format=1.3]|
}
\toprule
& \multicolumn{2}{c|}{\textbf{Summarization}} & \multicolumn{4}{c|}{\textbf{Bubble Detection}} & \multicolumn{4}{c|}{\textbf{Next-Page Pred.}} & \multicolumn{4}{c|}{\textbf{Inter-Page Pred.}} \\
\cmidrule(lr){2-3} \cmidrule(lr){4-7} \cmidrule(lr){8-11} \cmidrule(lr){12-15}
\textbf{Model} & {\textbf{TI}} & {\textbf{VI}} & {\textbf{IoU$\uparrow$}} & {\textbf{Prec$\uparrow$}} & {\textbf{Rec$\uparrow$}} & {\textbf{F1$\uparrow$}} & {\textbf{5p@1$\uparrow$}} & {\textbf{10p@1$\uparrow$}} & {\textbf{5pMRR$\uparrow$}} & {\textbf{10pMRR$\uparrow$}} & {\textbf{2p@1$\uparrow$}} & {\textbf{3p@1$\uparrow$}} & {\textbf{2pMRR$\uparrow$}} & {\textbf{3pMRR$\uparrow$}} \\
\midrule
InternVL3-2B  & 0.834 & 0.802 & 0.610 & \underline{\textbf{0.394}} & 0.218 & 0.281 &  4.0\% &  8.9\% & 0.350 & 0.378 & 13.6\% & 18.2\% & 0.285 & 0.332 \\
Ovis2-2B      & 0.833 & 0.811 & 0.580 & 0.065 & 0.054 & 0.059 & 40.9\% & 27.3\% & 0.581 & 0.477 & 22.7\% & 31.8\% & 0.398 & 0.477 \\
Qwen2.5-VL-3B & 0.837 & 0.810 & \textbf{0.630} & 0.088 & 0.051 & 0.064 & 20.0\% & 15.6\% & 0.479 & 0.411 & 15.9\% & 20.5\% & 0.318 & 0.367 \\
InternVL3-8B  & 0.835 & 0.808 & 0.609 & \textbf{0.472} & \underline{\textbf{0.255}} & \underline{\textbf{0.331}} & \underline{\textbf{43.2\%}} & \textbf{50.0\%} & \textbf{0.716} & \textbf{0.750} & \underline{\textbf{36.4\%}} & \underline{\textbf{43.2\%}} & \underline{\textbf{0.523}} & \underline{\textbf{0.614}} \\
Ovis2-8B      & 0.844 & \underline{\textbf{0.820}}& 0.608 & 0.057 & 0.049 & 0.053 & 35.5\% & 31.8\% & 0.531 & 0.477 & 25.0\% & 34.1\% & 0.432 & 0.523 \\
Qwen2.5-VL-7B & \underline{\textbf{0.845}} & 0.817 & \underline{\textbf{0.611}} & 0.221 & 0.263 & 0.240 & 25.0\% & 22.7\% & \underline{\textbf{0.544}} & \underline{\textbf{0.511}} & 18.2\% & 25.0\% & 0.356 & 0.432 \\
InternVL3-14B & 0.842 & \textbf{0.823} & 0.607 & 0.335 & \textbf{0.352} & \textbf{0.343} &  4.7\% & 13.6\% & 0.423 & 0.439 & 20.5\% & 27.3\% & 0.367 & 0.455 \\
Ovis2-16B     & \textbf{0.844} & \textbf{0.833} & 0.609 & 0.234 & 0.231 & 0.233 & \textbf{35.0\%} & \underline{\textbf{29.5\%}} & 0.567 & 0.523 & \textbf{29.5\%} & \textbf{36.4\%} & \textbf{0.477} & \textbf{0.550} \\
\bottomrule
\end{tabular}
\end{table*}

Table~\ref{tab:comprehensive-performance} reveals critical patterns across all evaluation dimensions. For visual-textual integration, all models achieve reasonable spatial localization (IoU 0.580-0.630) but suffer severe text extraction failures with F1 scores below 0.25. This \textit{visual-textual integration bottleneck} explains the visual processing penalties observed in summarization experiments. For temporal reasoning, InternVL3-8B shows remarkable improvement with longer context (43.2\%→50.0\% top-1 accuracy) while most models degrade, and the intuitive narrative constraint emerges where models consistently perform better on 3-missing-3 page scenarios than 2-missing-2 page ones.

\subsection{Temporal Reasoning: The Inferent Gap}

Our temporal reasoning experiments directly probe the inferent gap through next-page and intermediate-page prediction tasks. The results reveal architecture-dependent processing differences and counterintuitive patterns that illuminate fundamental challenges in narrative understanding.

\begin{figure}[t]
\centering
\begin{subfigure}[t]{\linewidth}
    \centering
    \includegraphics[width=\linewidth]{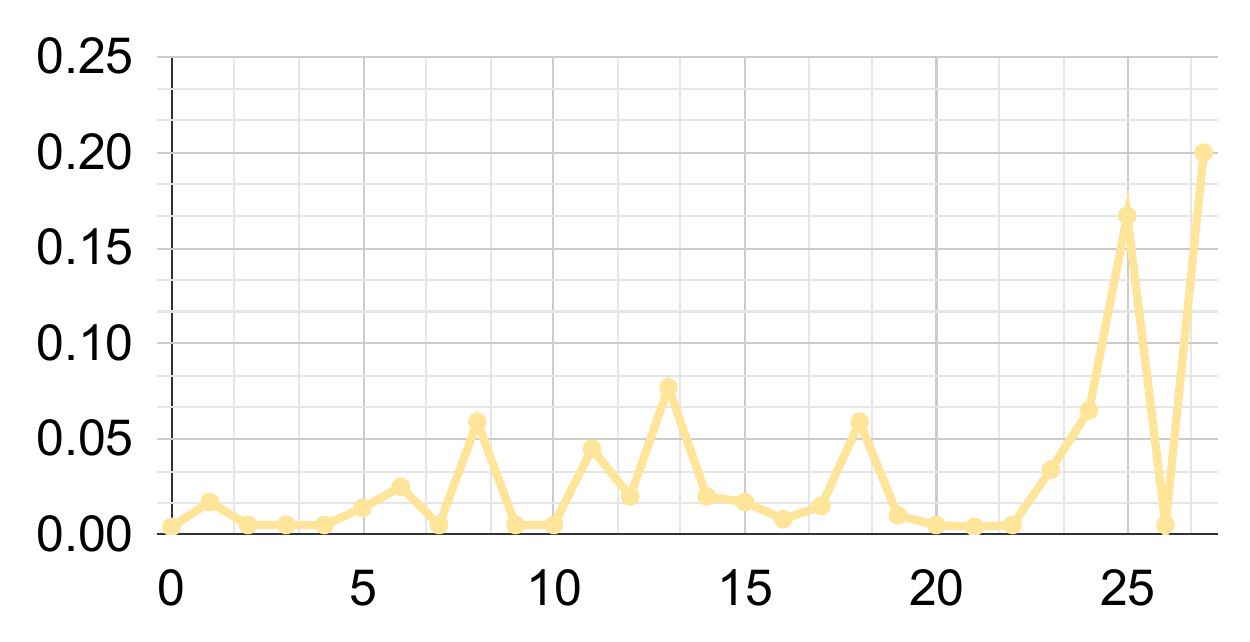}
    \caption{\textbf{In-Chapter Analysis}}
    \label{fig:chapter7_detail}
\end{subfigure}

\vspace{1em}
\begin{subfigure}[t]{\linewidth}
    \centering
    \includegraphics[width=\linewidth]{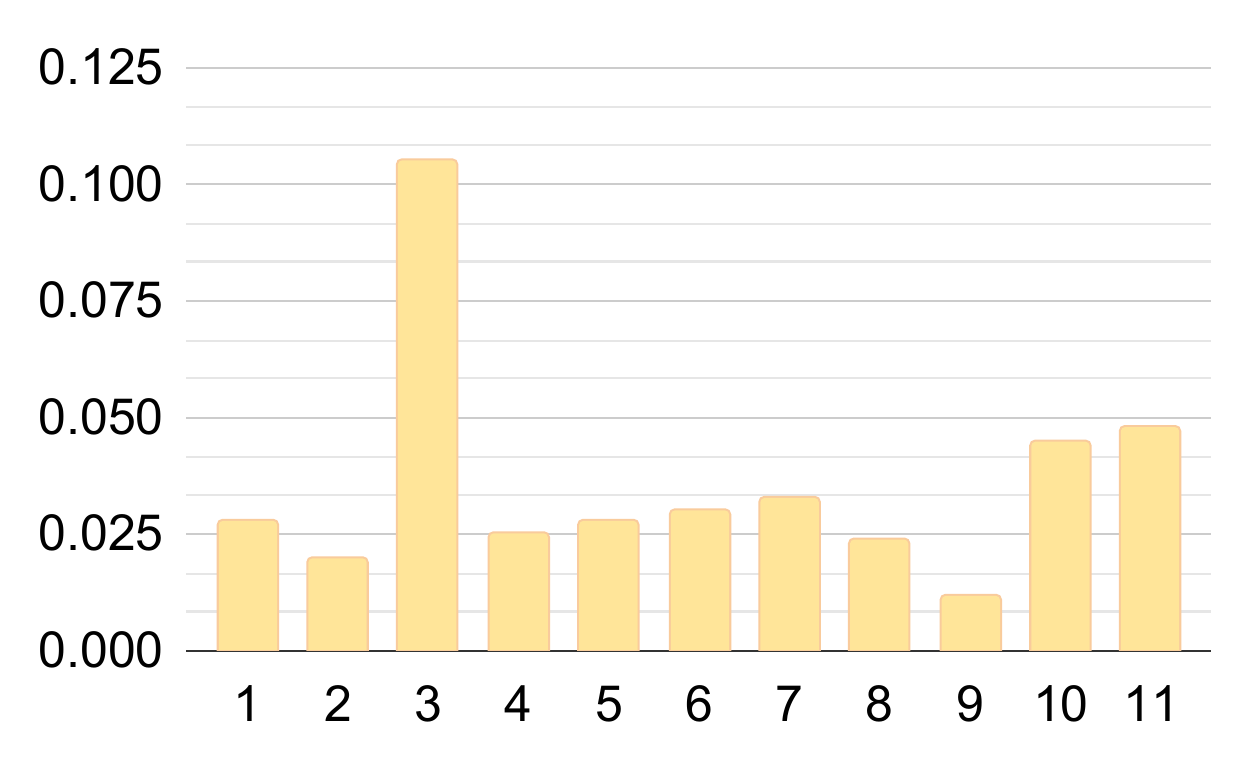}
    \caption{\textbf{Over-Chapter Analysis}}
    \label{fig:mrr_overall}
\end{subfigure}
\caption{
\textbf{\subref{fig:chapter7_detail}:} Mean Reciprocal Rank (MRR) across all chapters (x-asis: chapter number; y-axis: average MMR Scores), highlighting the cosistent overall poor scores on retrieving the visual counterpart provided the text.
\textbf{\subref{fig:mrr_overall}:}  Mean Reciprocal Rank (MRR) across pages of Chapter 7 (x-asis: page index; y-axis: MMR Score), highlighting the poor scores on retrieving the visual counterpart provided the text within the confines of a specific chapter.
}
\label{fig:mrr_performance_analysis}
\end{figure}

\textbf{Architecture-Dependent Processing Differences.} Our analysis reveals that architecture design impacts comics understanding more significantly than parameter count. The InternVL3-8B model's superior performance over larger models from other architectures suggests specialized temporal processing mechanisms are crucial. Notably, the Ovis2 series demonstrates dramatic scale-dependent improvement (16B model shows only 1.1 point cross-modal penalty vs. 2.2 points for 2B model), while Qwen2.5-VL maintains consistent 2.7-2.8 point penalties across scales, indicating architecture-specific scaling behaviors for visual narrative processing, even in non-performant scores.

\subsection{Semantic Discontinuity in Narratives}

Our retrieval experiments using four vision encoders (BLIP, CLIP, SIGLIP, ALIGN) provide quantitative evidence for semantic discontinuity in comics. Chapter-wise similarity analysis reveals highly variable semantic patterns demonstrating the inferent gap within and over the chapters.

\begin{figure}[t]
\centering
\includegraphics[width=\linewidth]{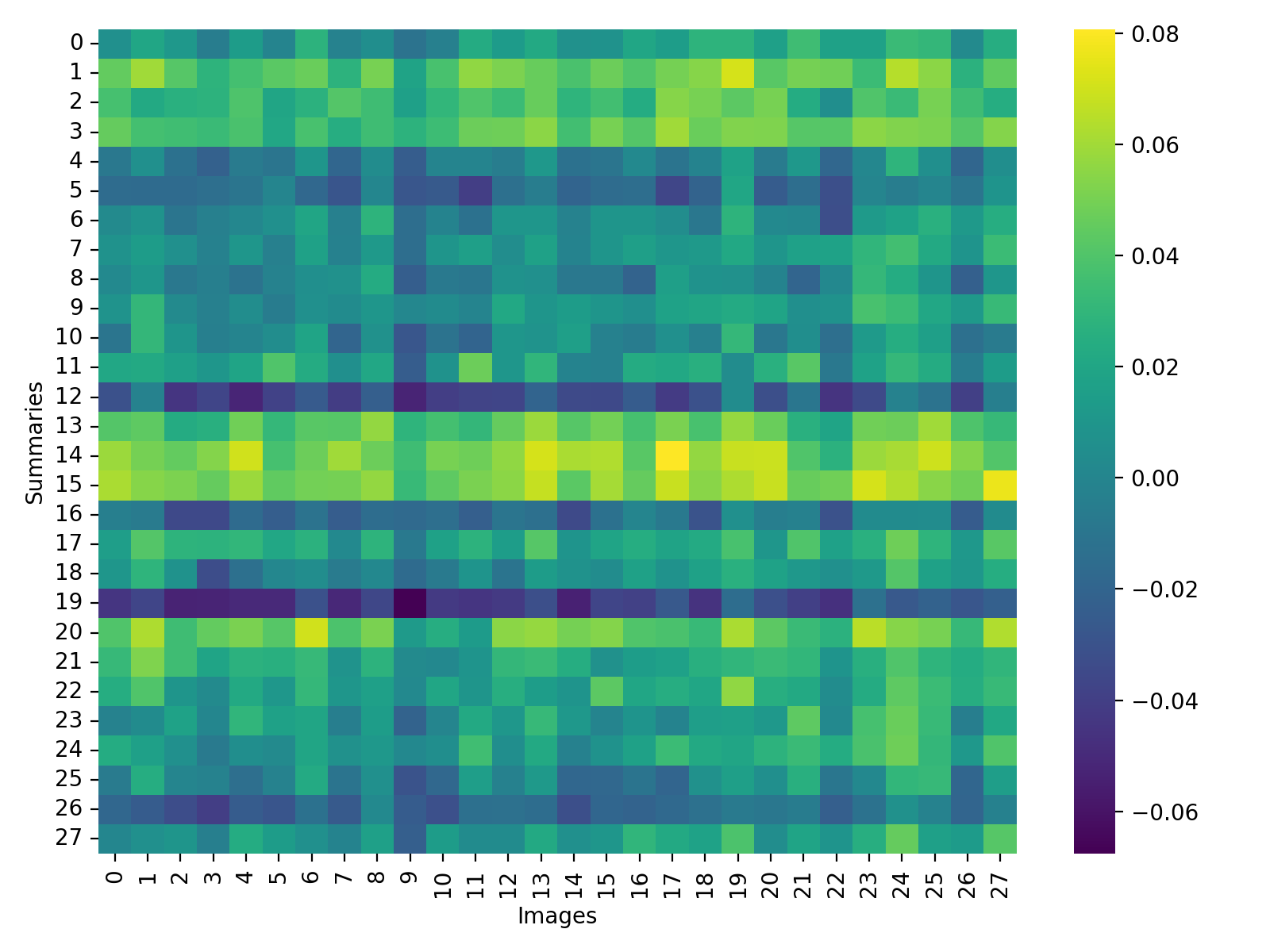}
\caption{Chapter 7 similarity heatmap demonstrating semantic discontinuity patterns between manga pages and corresponding text using SIGLIP encoder. Irregular patterns reveal the inferent gap challenge in sequential narrative understanding.}
\label{fig:heatmap_chapter_007}
\end{figure}

Figure~\ref{fig:heatmap_chapter_007} shows that vision encoders struggle to maintain semantic continuity between consecutive pages. SIGLIP similarity scores range from -0.088 to 0.230, with negative scores indicating semantically orthogonal sequential pages. Volume-wide retrieval performance is consistently poor across all embedding methods (0.047-0.076 normalized similarity). CLIP achieves the highest performance at 0.076, representing less than 8\% of theoretical maximum.

\textbf{Chapter-Specific Embedding Patterns.} Our analysis reveals systematic patterns in semantic discontinuity: early chapters (1-3) show highest semantic discontinuity with similarity ranges of 0.1-0.3 across all encoders, particularly challenging during character introduction sequences; mid-chapters (4-7) demonstrate moderate discontinuity (0.2-0.4 ranges) during plot development; while late chapters (8-11) achieve best coherence (0.3-0.5 ranges) in action-oriented sequences, suggesting current vision encoders are more amenable to visual action than narrative dialogue.

\subsection{VQA: Temporal Reasoning Degradation}

Our comprehensive Visual Question Answering evaluation across all 11 chapters provides direct evidence of systematic temporal reasoning failures (Table~\ref{tab:vqa_chapter_performance}). The results demonstrate severe temporal reasoning limitations, with overall performance averaging only 1.43/5.0 (28.5\% accuracy).

\begin{table}[t]
\centering
\caption{\textbf{VQA performance across narrative progression stages (0–5 scale).} \textit{Later chapters reveal consistent degradation in reasoning ability. Bold-underline marks top 2 models overall.}}
\footnotesize
\resizebox{\linewidth}{!}{%
\begin{tabular}{|l|cccccc|c|}
\toprule
\textbf{Model} & \textbf{Ch 1-2\up} & \textbf{Ch 3-4\up} & \textbf{Ch 5-6\up} & \textbf{Ch 7-8\up} & \textbf{Ch 9-10\up} & \textbf{Ch 11\up} & \textbf{Overall\up} \\
\midrule

InternVL3-2B      & 1.50 & 1.50 & 1.30 & 1.10 & 1.20 & 1.00 & 1.29 \\
Ovis2-2B          & 2.00 & 1.50 & 1.30 & 1.40 & 1.00 & 1.20 & 1.42 \\
Qwen2.5-VL-3B     & 1.80 & 1.30 & 1.60 & 1.00 & 1.20 & 1.20 & 1.36 \\
InternVL3-8B      & 1.20 & 1.30 & 1.70 & 1.50 & 1.40 & 0.80 & 1.36 \\
Ovis2-8B          & 1.80 & 1.60 & 1.40 & 1.40 & 1.50 & 0.80 & 1.47 \\
Qwen2.5-VL-7B     & \textbf{2.10} & 1.20 & 1.30 & \textbf{1.80} & \textbf{2.50} & 1.00 & \underline{\textbf{1.71}} \\
InternVL3-14B     & 1.90 & 1.40 & 1.50 & 1.60 & 1.45 & 0.80 & \textbf{1.50} \\
Ovis2-16B         & 1.80 & 1.50 & 1.50 & 1.50 & 1.20 & 0.80 & 1.44 \\
\bottomrule
\end{tabular}
}
\label{tab:vqa_chapter_performance}
\end{table}

The results demonstrate severe temporal reasoning limitations, with overall performance averaging only 1.43/5.0 (28.5\% accuracy). Critically, 7/9 models show performance degradation from early to late chapters (0.20-0.65 point drops). Chapter 11's narrative climax proves most challenging, with 8/9 models achieving lowest performance (0.80-1.20/5.0). This systematic temporal degradation provides direct evidence that VLMs struggle as narrative complexity accumulates across discrete visual sequences. High chapter-wise variance (standard deviations 0.207-0.616) reveals inconsistent temporal understanding, indicating models process pages independently.

\textbf{RAG Enhancement Analysis.} To investigate whether external knowledge retrieval could mitigate these limitations, we evaluated models with and without retrieval-augmented generation (RAG). While RAG reduces complete failures (score 0) from ~70\% to ~40\% for models like Qwen2.5-VL-7B, high-quality responses (scores 4-5) remain extremely rare across all models. This demonstrates that temporal reasoning failures persist even with external knowledge augmentation, confirming that the core limitations stem from architectural inability to process discrete temporal structures rather than knowledge gaps.

These findings demonstrate that current VLMs cannot effectively process the discrete temporal modality defining comics as a unique narrative medium.

\subsection{Novel Narrative Evaluation Methodology}

To comprehensively assess VLM narrative generation quality, we introduce a \textbf{dual-framework evaluation system} building upon prior work (SCORE~\cite{yi2025scorestorycoherenceretrieval}, AIStorySimilarity~\cite{chun2024aistorysimilarity}) but with several key innovations. Unlike binary or pass/fail approaches, we introduce proportional penalty systems, minimum score thresholds, and state-based incremental evaluation, enabling fine-grained differentiation of narrative quality. We systematically evaluate four core narrative dimensions on a 0-100 scale: Character Development, Plot Structure, Setting/Atmosphere, and Thematic Coherence. Score ranges provide clear quality differentiation from 90-100 (Exceptional) to 0-9 (Completely inadequate), enabling fine-grained comparison between models and revealing performance gradients invisible to binary evaluation systems. This represents the first penalty and score-based state metric for narrative quality assessment.

\subsection{Narrative Coherence Breakdown}

We conducted comprehensive narrative analysis using SCORE (Story Coherence and Retrieval Enhancement) framework and strict narrative similarity analysis. Table~\ref{tab:narrative_evaluation} presents OVIS2-16B evaluation results across all three tasks using our brutal narrative analyzer with penalties for narrative inconsistencies as parameterized in the framework.

\begin{table}[t]
\centering
\footnotesize
\caption{\textbf{Best narrative evaluation achieved by Ovis2-16B.} Higher values indicate better performance except for StdDev (↓).}
\label{tab:narrative_evaluation}
\resizebox{\linewidth}{!}{%
\begin{tabular}{|l|ccccccc|}
\toprule
\textbf{Task} & \textbf{Overall↑} & \textbf{Character↑} & \textbf{Plot↑} & \textbf{Setting↑} & \textbf{Best↑} & \textbf{Worst↑} & \textbf{StdDev↓} \\
\midrule
Task 1 & 28.36 & 31.27 & 27.27 & 20.64 & \textbf{31.25} & 19.50 & 3.44 \\
Task 2 & 28.34 & \textbf{32.00} & \textbf{28.00} & 19.09 & \textbf{31.25} & \textbf{27.50} & \textbf{1.53} \\
Task 3 & \textbf{28.70} & \textbf{32.00} & \textbf{28.00} & \textbf{21.64} & \textbf{31.25} & 25.00 & 2.44 \\
\midrule
\textbf{Average} & \textbf{28.47} & \textbf{31.76} & \textbf{27.76} & \textbf{20.46} & \textbf{31.25} & \textbf{24.00} & \textbf{2.47} \\
\bottomrule
\end{tabular}
}
\end{table}

Results reveal systematic narrative failures across all dimensions. With average overall score of 28.47/100, OVIS2-16B demonstrates fundamental limitations in constructing coherent long-form narratives. Character development (31.76/100) marginally outperforms plot construction (27.76/100), while setting development shows severe deficiencies (20.46/100). These findings confirm the \textit{discrete temporal processing bottleneck} identified earlier. SCORE framework evaluation reveals VLMs process manga pages as isolated units rather than unified temporal narrative elements. Consistently low scores indicate current VLMs lack architectural components for maintaining narrative coherence across extended sequential contexts.

\textbf{Narrative Depth Penalties.} Our strict analyzer applies harsh penalties for insufficient story development. Observed scores reflect both poor narrative quality and fundamental inability to generate content with sufficient depth for meaningful story construction. This aligns with character attribution findings, suggesting VLMs struggle with hierarchical binding across multiple temporal contexts---a core manga understanding requirement.

%% file: sec/5_conclusion.tex
\section{Conclusion}

We present Re:Verse, a comprehensive benchmark for evaluating VLMs on manga narrative understanding. Through systematic analysis across story synthesis, character grounding, and temporal reasoning, we demonstrate that current VLMs face fundamental challenges in processing the discrete temporal structure of visual narratives.

\subsection{Key Contributions}

\textbf{Novel Benchmark Design.} We introduce Re:Verse, the first benchmark for chapter-length manga narrative understanding. Unlike existing benchmarks focusing on isolated panels, Re:Verse features 308 annotated pages from 11 chapters with aligned narrative text, enabling systematic evaluation of long-form story comprehension. 

\textbf{Comprehensive Evaluation.} We conduct extensive evaluation of 8 SOTA opensource VLMs across story synthesis, visual-textual integration, and temporal reasoning. 

\textbf{Identification of Fundamental Limitations.} We identify three critical bottlenecks: (i) character consistency failures with 3-10x gaps in named entity recognition, (ii) visual-textual integration bottlenecks with F1 scores below 0.34, and (iii) temporal reasoning degradation with 28.5\% accuracy on sequential understanding tasks.

\textbf{Novel Evaluation Methodology.} We introduce the first penalty and score-based state metric for narrative quality assessment, building upon SCORE~\cite{yi2025scorestorycoherenceretrieval} and AIStorySimilarity~\cite{chun2024aistorysimilarity} frameworks. Our dual-framework system provides quantitative differentiation between poor-quality outputs rather than binary assessment.

\textbf{Quantification of the Inferent Gap.} First quantitative evidence for the \textit{inferent gap} problem through intermediate-page prediction, showing models paradoxically perform better on 3-missing-page than 2-missing-page scenarios.

\subsection{Impact and Future Directions}

The systematic temporal reasoning degradation demonstrates models lack discrete temporal processing mechanisms necessary for narrative understanding. The Re:Verse benchmark provides a foundation for future research in sequential visual narrative understanding, extending beyond comics to other visual storytelling forms.

\textbf{Broader Impact.} This work represents a significant step towards enabling contextual understanding of manga and comics---culturally significant visual media that have been underserved by the vision community. Our benchmark enables development of context-aware applications for digital humanities research, creative assistance, and educational tools that may meaningfully engage with visual narratives.

\textit{* Re:Verse will be distributed under the CC BY-NC 4.0 license, permitting use for strictly non-commercial purposes. Access will be conditioned on explicit and mandatory acceptance of the license terms via a permission gate.}

%% file: sec/6_limitations.tex
\section{Limitations}

While Re:Verse establishes a new frontier, primary constraint is the scarcity of source material, given the high degree of narrative alignment between the \textit{Re:Zero} manga and its light novel, the labor-intensive process required to create a high-fidelity dataset, ensures quality but limits its scale.

%% file: sec/X_suppl.tex
\appendix
\clearpage
\setcounter{page}{1}

\section*{Supplementary Material}
\label{sec:detailed_analysis}

This supplementary material provides comprehensive chapter-by-chapter analysis of VLM performance on manga narrative understanding, with detailed results for all 11 chapters including semantic similarity heatmaps. Core RAG enhancement findings and architecture-specific processing differences are discussed in the main paper (Section 4.4-4.5), while this supplement focuses on granular chapter-wise breakdowns and extended technical details.

\section*{Extended Evaluation Methodology}
\label{sec:eval_methodology}

This section provides additional technical specifications for our dual-framework evaluation system beyond those detailed in Section 3.3 of the main paper, focusing on advanced implementation details and computational considerations.

\subsection*{Proportional Penalty System}

Unlike traditional binary evaluation approaches that assign zero scores to inadequate content, our proportional penalty system applies graduated multipliers based on content quality assessment:

\paragraph{Severity-Based Multipliers:} Content quality is assessed across four dimensions (Character Development, Plot Structure, Setting/Atmosphere, Thematic Coherence) with multipliers ranging from 0.6x to 0.9x applied based on deficiency severity:
\begin{itemize}
\item \textbf{Minor deficiencies (0.9x multiplier):} Superficial issues that don't fundamentally compromise narrative understanding
\item \textbf{Moderate deficiencies (0.8x multiplier):} Noticeable problems that impact but don't destroy narrative coherence  
\item \textbf{Major deficiencies (0.7x multiplier):} Significant structural issues that severely compromise story quality
\item \textbf{Critical deficiencies (0.6x multiplier):} Fundamental failures that nearly eliminate narrative value
\end{itemize}

\paragraph{Composite Scoring:} Final scores are calculated using weighted aggregation: $S_{final} = \sum_{i=1}^{4} w_i \times S_{base,i} \times M_i$ where $w_i$ represents dimension weights, $S_{base,i}$ is the base score for dimension $i$, and $M_i$ is the corresponding severity multiplier.

\subsection*{Minimum Score Thresholds}

To ensure meaningful differentiation even among poor-quality outputs, we enforce dimension-specific minimum score thresholds:

\begin{itemize}
\item \textbf{Character Development:} 3/100 minimum (ensures basic character presence recognition)
\item \textbf{Plot Structure:} 2/100 minimum (acknowledges any narrative sequence attempt)
\item \textbf{Setting/Atmosphere:} 4/100 minimum (recognizes environmental context awareness)  
\item \textbf{Thematic Coherence:} 1/100 minimum (credits any thematic element recognition)
\end{itemize}

These thresholds prevent total score collapse while maintaining evaluative rigor, enabling fine-grained comparison between fundamentally flawed but differently deficient outputs.

\subsection*{State-Based Incremental Evaluation}

Our evaluation system implements persistent state management to enable large-scale assessment across model families:

\paragraph{Incremental Processing:} Results are saved after each individual evaluation, preventing data loss during extended evaluation sessions spanning multiple days.

\paragraph{Resumable Assessment:} The system maintains evaluation state, allowing researchers to pause and resume comprehensive model comparisons without losing progress.

\paragraph{Metadata Preservation:} Complete evaluation context (model configurations, prompt variations, timestamp information) is preserved with each result for full reproducibility.

\subsection*{Quality Differentiation Framework}

Our 0-100 scale provides granular quality assessment with clear interpretation guidelines:

\begin{itemize}
\item \textbf{90-100 (Exceptional):} Publication-ready quality with minimal revision needs
\item \textbf{70-89 (Good):} Strong foundation requiring minor improvements  
\item \textbf{50-69 (Acceptable):} Adequate structure needing significant enhancement
\item \textbf{30-49 (Poor):} Fundamental structural issues requiring major revision
\item \textbf{10-29 (Severely Inadequate):} Critical failures needing complete reconstruction
\item \textbf{0-9 (Completely Inadequate):} Failed generation with no recoverable elements
\end{itemize}

This framework enables researchers to identify specific improvement areas and track incremental progress in model development, providing actionable insights beyond binary pass/fail assessment.

\section*{Extended RAG vs Non-RAG Analysis}
\label{sec:rag_analysis}

This section provides detailed distribution analysis and model-specific patterns for RAG enhancement, building upon the core findings presented in Section 4.4 of the main paper. The following figures show comprehensive score distributions across the 0-5 scale for all evaluated models.

\begin{figure*}[h]
\centering
\includegraphics[width=\textwidth]{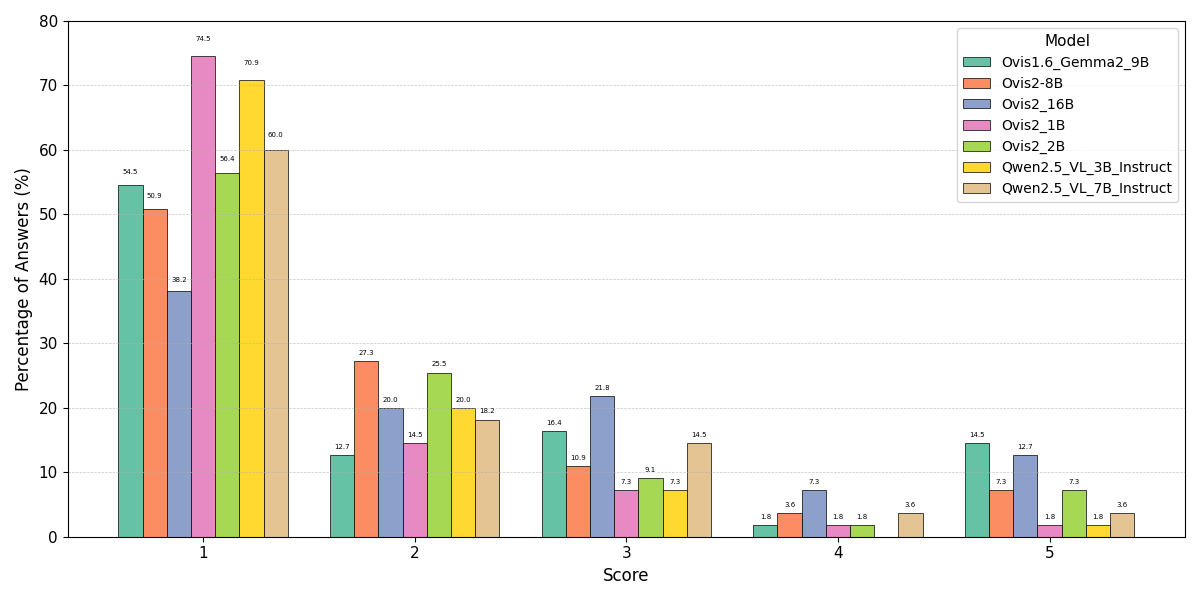}
\caption{Score distribution for VQA evaluation without RAG enhancement. The distribution shows heavy concentration in lower scores (0-2), with most models achieving predominantly poor performance. Notable patterns include high failure rates (score 0) across all models and minimal achievement of higher scores (4-5).}
\label{fig:without_rag}
\end{figure*}

\begin{figure*}[h]
\centering
\includegraphics[width=\textwidth]{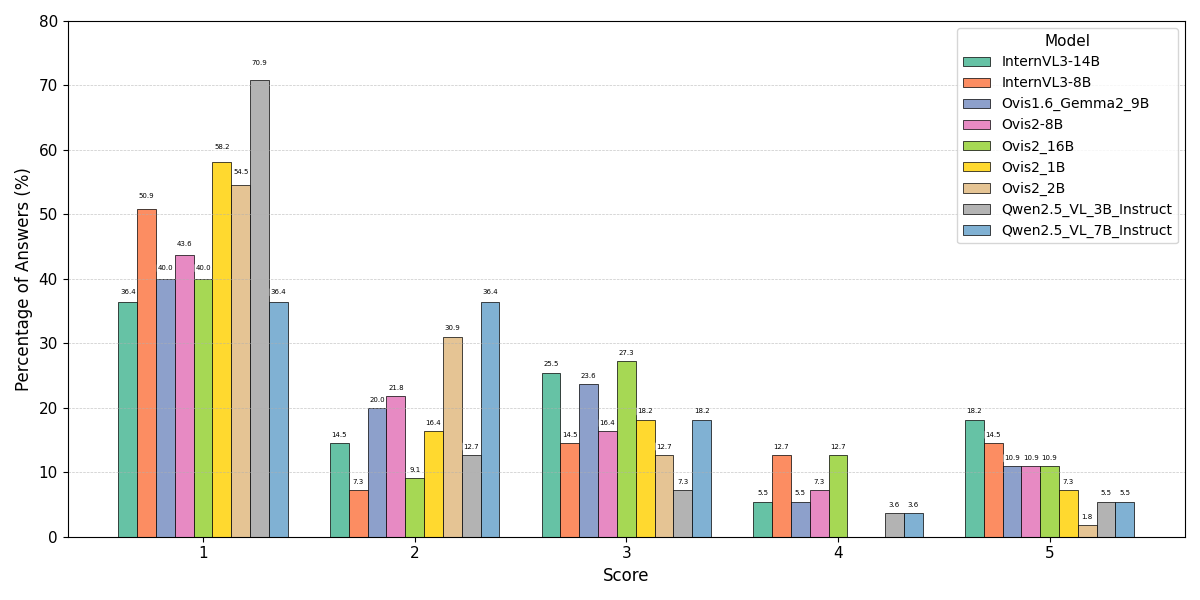}
\caption{Score distribution for VQA evaluation with RAG enhancement. The distribution shows improved performance with reduced failure rates and better score distribution across the 1-3 range. However, high-quality responses (scores 4-5) remain extremely rare, indicating fundamental limitations persist even with retrieval augmentation.}
\label{fig:with_rag}
\end{figure*}

\subsection*{RAG Impact Analysis}

The comparison between Figures~\ref{fig:without_rag} and \ref{fig:with_rag} reveals several critical insights beyond those summarized in the main paper:

\paragraph{Model-Specific RAG Benefits:} Different architectures show varying degrees of improvement with RAG. InternVL3 models demonstrate more consistent improvement patterns, while Ovis2 models show more variable responses to retrieval augmentation. Detailed per-model analysis shows Qwen2.5-VL-7B achieves the most dramatic improvement, while smaller models show minimal RAG benefits.

\paragraph{Score Distribution Granularity:} RAG implementation shifts the score distribution toward moderate performance levels (scores 1-2), with specific patterns varying by architecture family. The granular distribution analysis reveals that while RAG reduces catastrophic failures, it cannot elevate performance to high-quality ranges across any architecture.

\subsection*{Implications for Temporal Reasoning}

The extended RAG analysis provides additional evidence for our core thesis about discrete temporal processing limitations discussed in Section 4:

\paragraph{Retrieval Cannot Substitute Temporal Understanding:} While RAG improves response coherence, detailed analysis confirms it cannot compensate for the fundamental inability to process temporal dependencies across discrete visual sequences, as demonstrated in the main paper's architecture-specific findings.

\paragraph{Computational Limitations vs. Knowledge Gaps:} The results provide definitive evidence that manga understanding challenges stem from computational architecture limitations rather than knowledge deficiencies, supporting the main paper's conclusions about the need for specialized temporal processing mechanisms.

\subsection*{Individual Chapter Performance Analysis}

The chapter-wise analysis provides granular evidence for the semantic discontinuity patterns and architecture-specific behaviors discussed in Section 4.3 of the main paper. The following detailed breakdowns illuminate specific failure modes across different narrative contexts.

\subsubsection*{Early Chapters (1-3): Character Introduction}

The initial chapters focus on character introduction and world-building, exemplifying the semantic discontinuity patterns identified in the main paper (similarity ranges 0.1-0.3 across all encoders).

\begin{figure*}[h]
\centering
\includegraphics[width=0.65\linewidth]{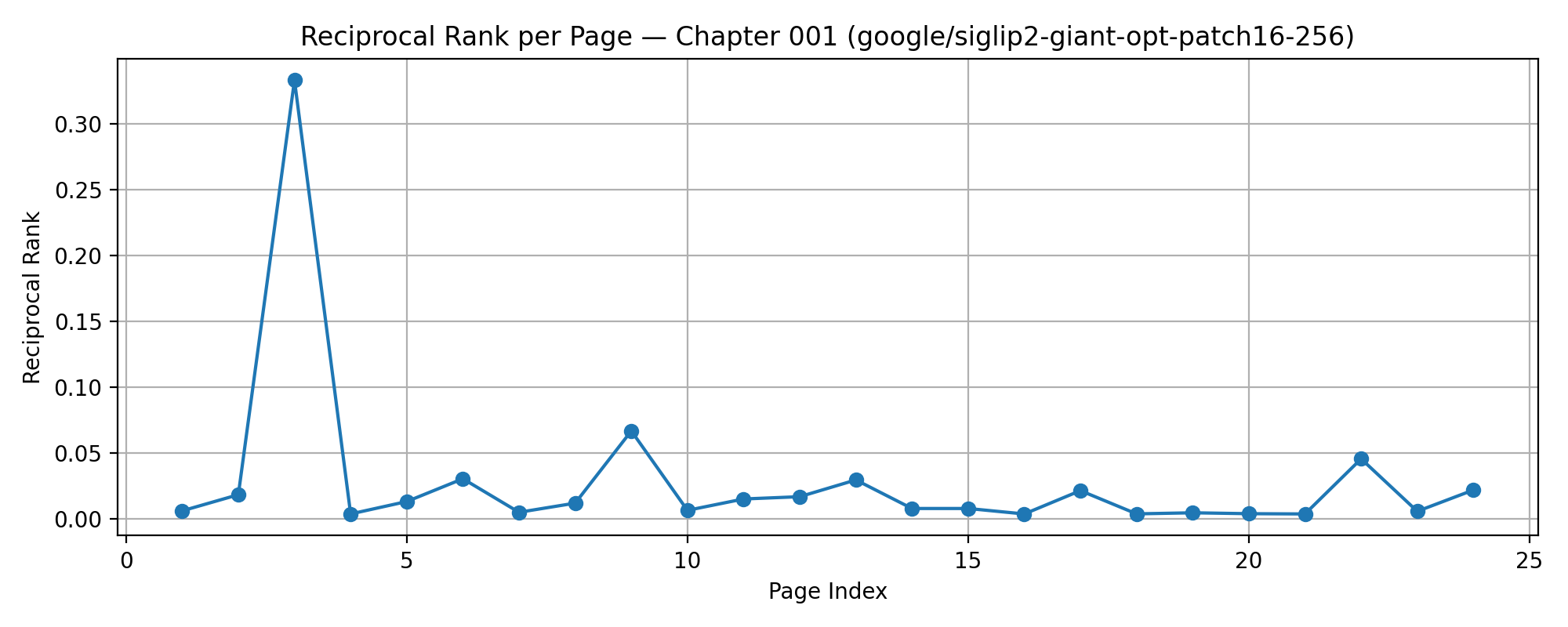}
\hfill
\includegraphics[width=0.34\linewidth]{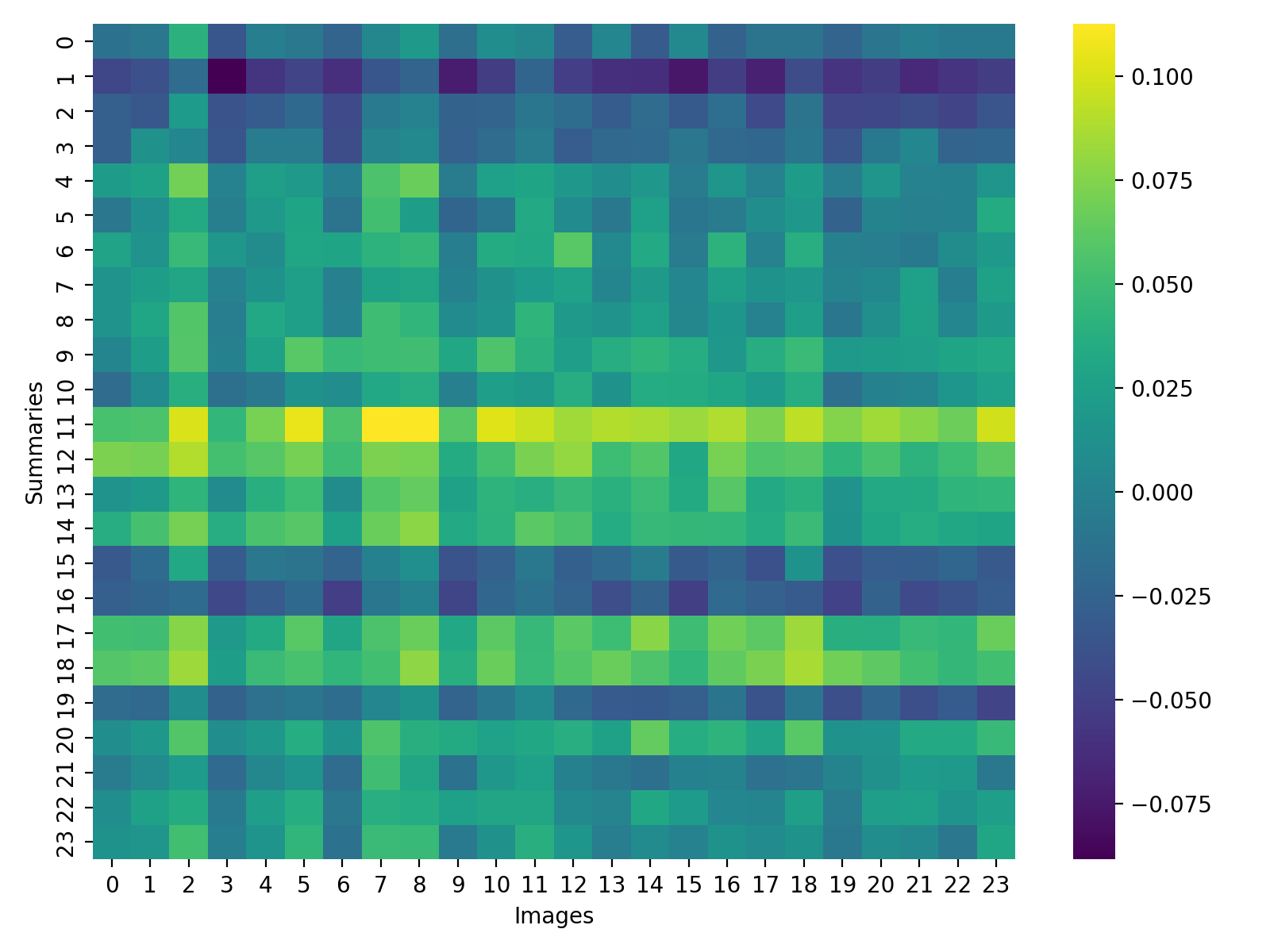}
\caption{Chapter 1 performance analysis (left) and semantic similarity heatmap (right). The heatmap shows irregular patterns with low inter-page similarity, indicating challenges in maintaining narrative coherence during character introduction sequences. 
\SV{make other figures page width too. see if you can increase font size for labels etc, not readable. confusion matrix do not require axis labels, say what it is in caption, }
}
\label{fig:chapter_001}
\end{figure*}

\begin{figure}[h]
\centering
\includegraphics[width=0.48\linewidth]{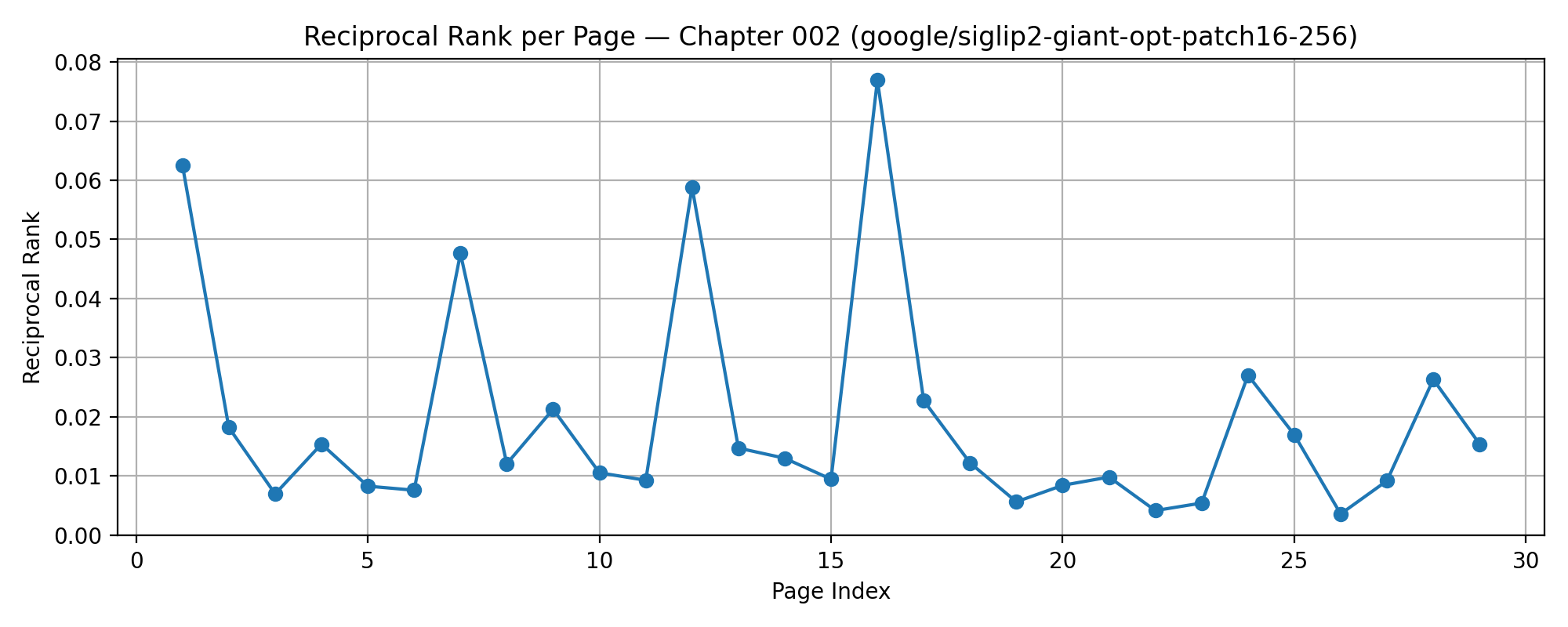}
\hfill
\includegraphics[width=0.48\linewidth]{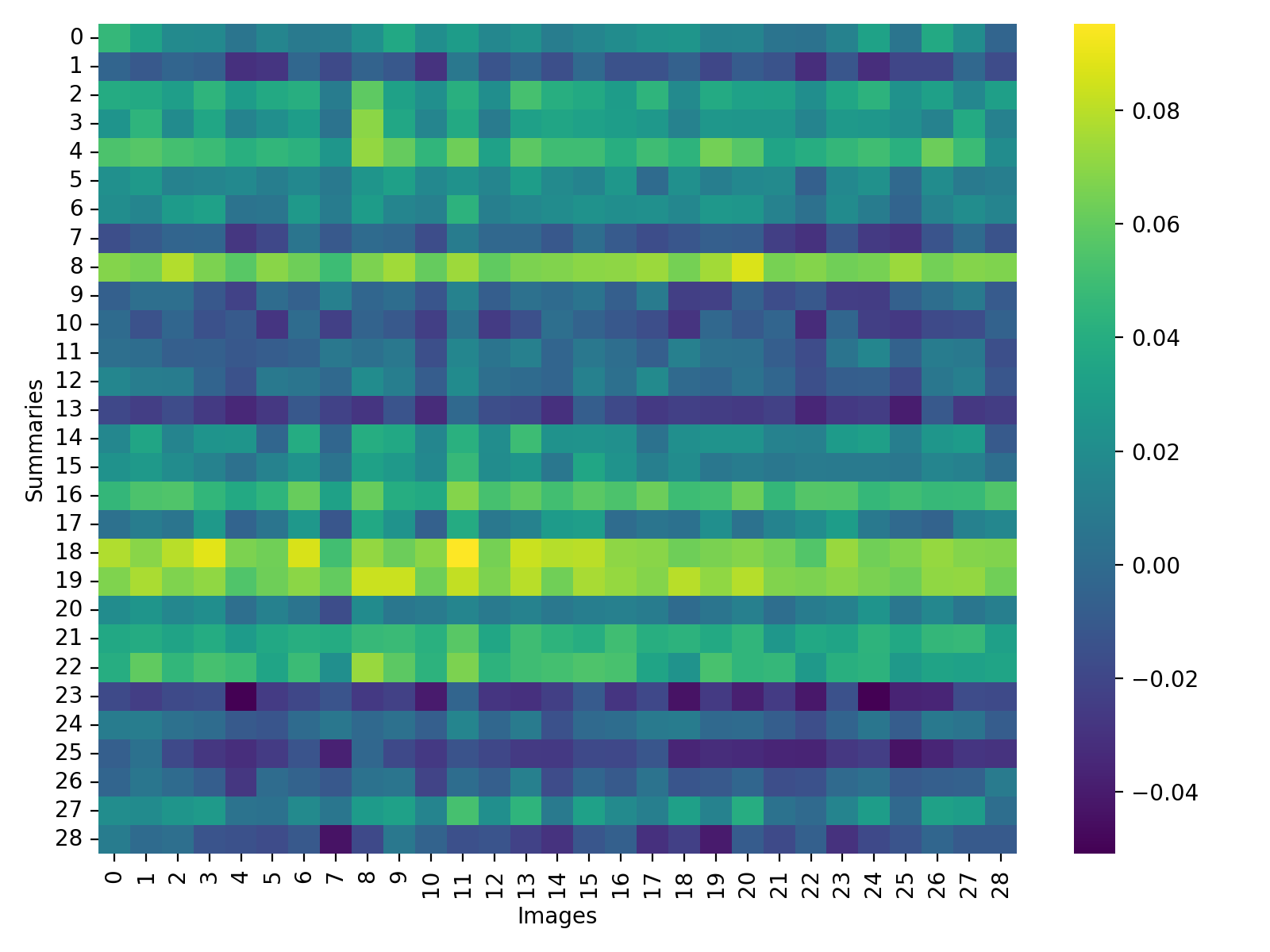}
\caption{Chapter 2 performance analysis (left) and semantic similarity heatmap (right). The extended length of Chapter 2 (28 pages) creates additional challenges for maintaining temporal coherence across longer sequences.}
\label{fig:chapter_002}
\end{figure}

\begin{figure}[h]
\centering
\includegraphics[width=0.48\linewidth]{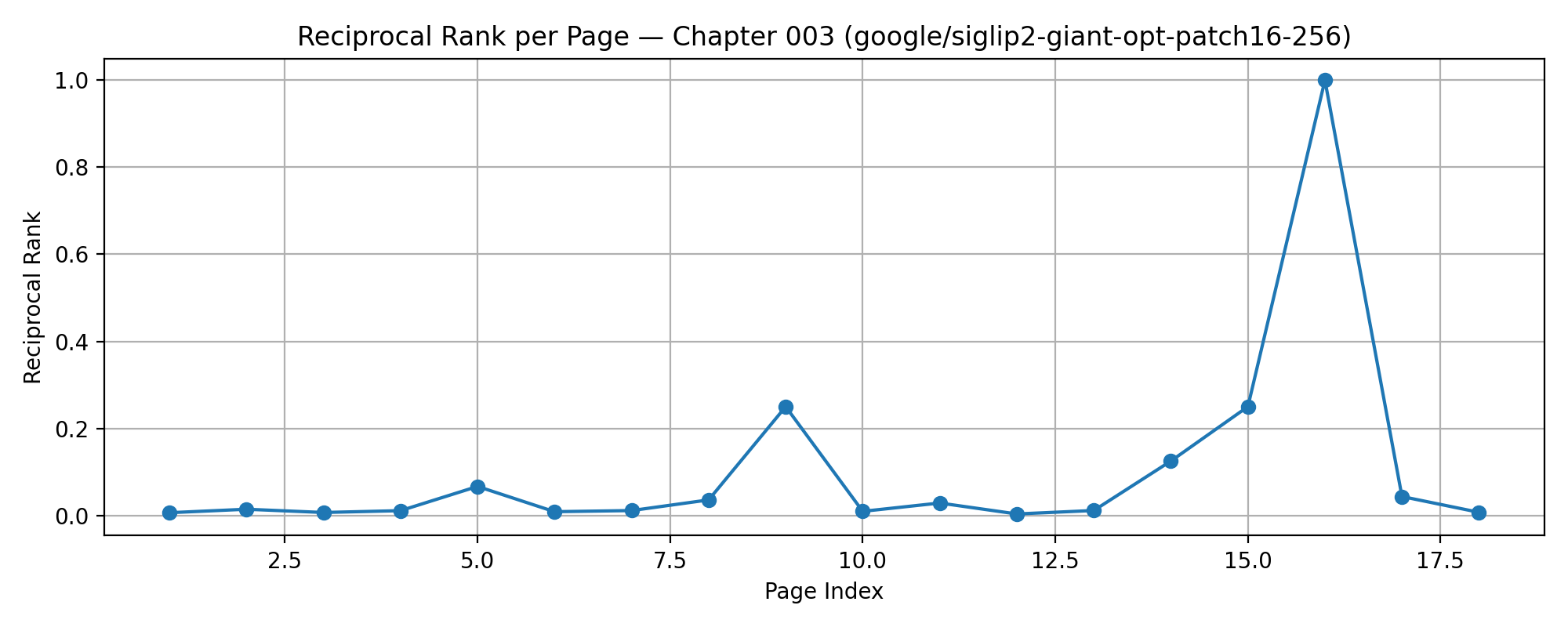}
\hfill
\includegraphics[width=0.48\linewidth]{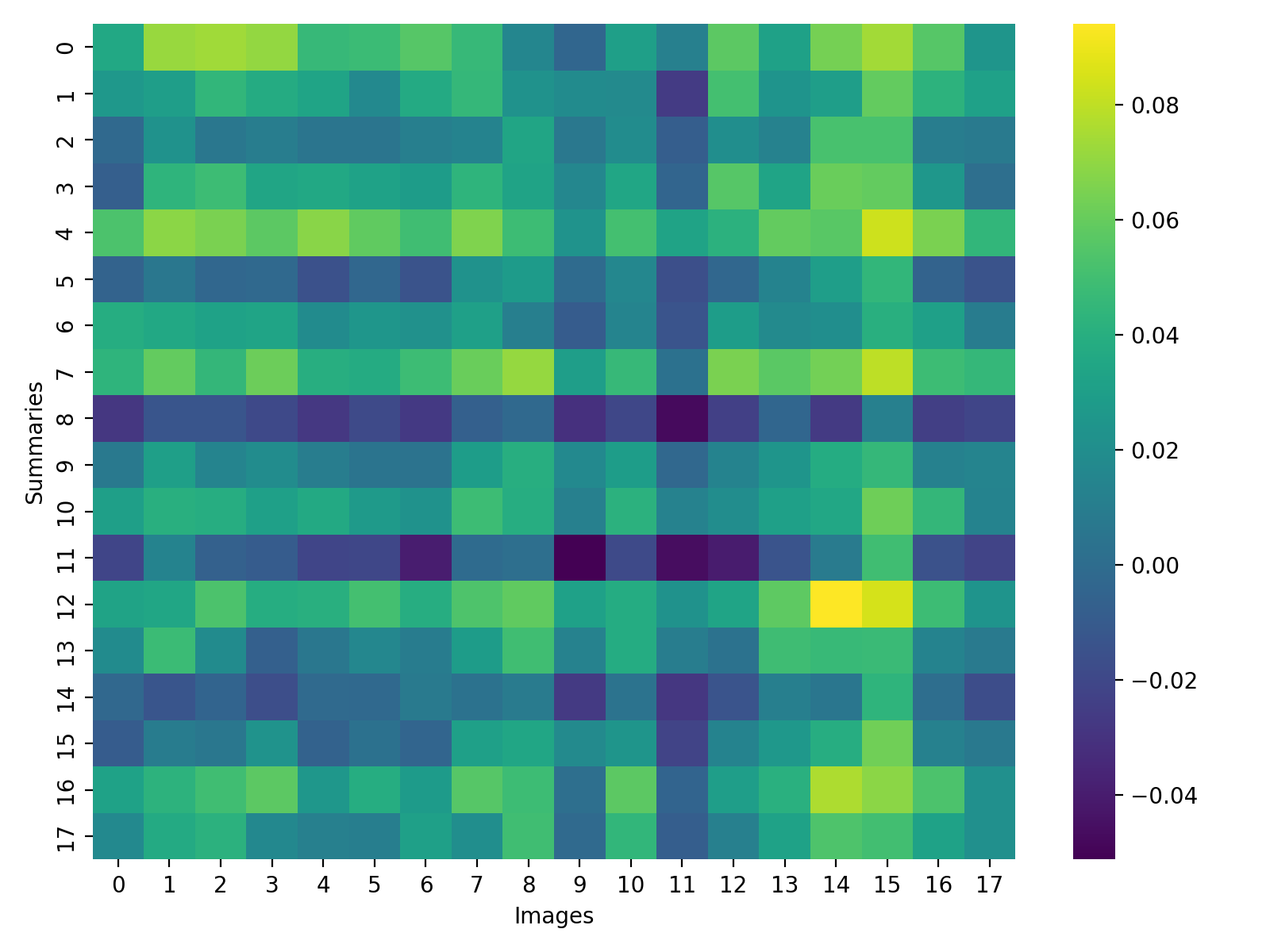}
\caption{Chapter 3 performance analysis (left) and semantic similarity heatmap (right). The heatmap reveals patches of higher similarity, suggesting some narrative segments are more coherent than others.}
\label{fig:chapter_003}
\end{figure}

Chapter 1 shows particularly challenging patterns, with similarity scores ranging from near-zero to moderate values (0.1-0.2), indicating that even consecutive pages often lack semantic continuity as captured by current vision encoders. The performance analysis reveals that all models struggle with the character introduction sequences, with NER density scores consistently below 0.03 compared to the gold standard of 0.087.

Chapter 2, being the longest chapter with 28 pages, presents additional challenges for temporal reasoning tasks. The similarity heatmap shows fragmented patterns with only sparse clusters of high similarity, suggesting that current vision encoders cannot effectively capture the narrative flow across extended sequences.

Chapter 3 shows some improvement in semantic coherence, with the heatmap revealing diagonal patterns that indicate better sequential understanding. However, the performance metrics still show significant gaps compared to human-level understanding.

\subsubsection*{Mid-Chapters (4-7): Plot Development Complexity}

The middle chapters focus on plot development and character relationships, presenting complex narrative structures that test the limits of current VLMs.

\begin{figure}[h]
\centering
\includegraphics[width=0.48\linewidth]{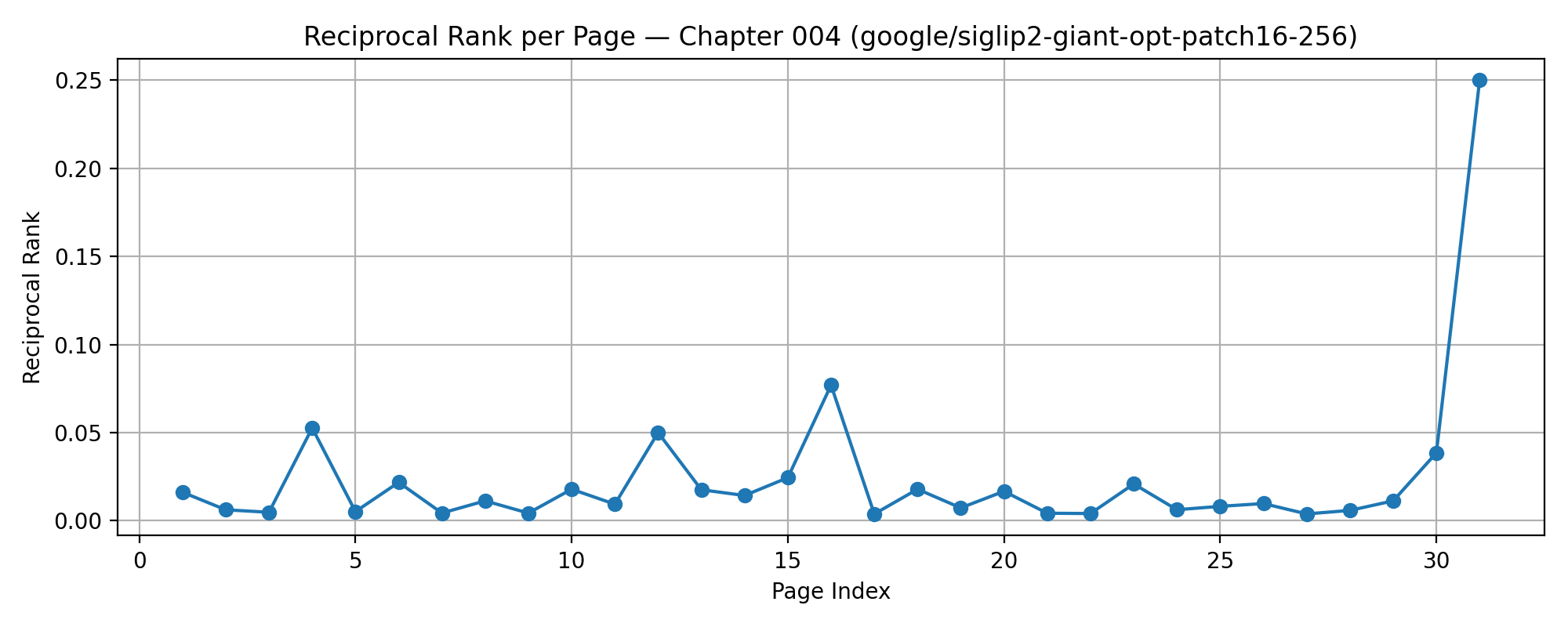}
\hfill
\includegraphics[width=0.48\linewidth]{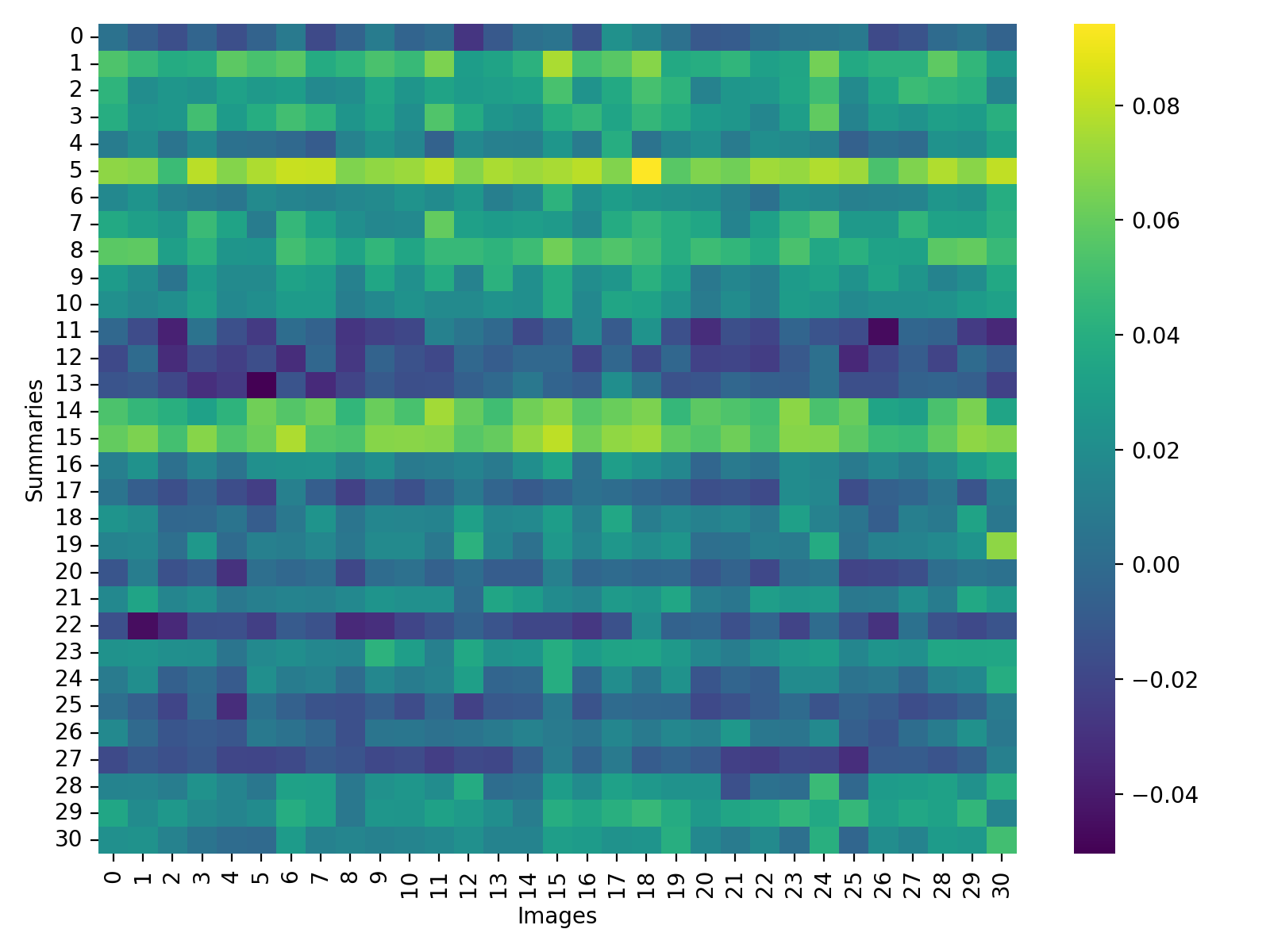}
\caption{Chapter 4 performance analysis (left) and semantic similarity heatmap (right). The complex plot development in Chapter 4 creates challenging inference requirements that all models struggle with.}
\label{fig:chapter_004}
\end{figure}

\begin{figure}[h]
\centering
\includegraphics[width=0.48\linewidth]{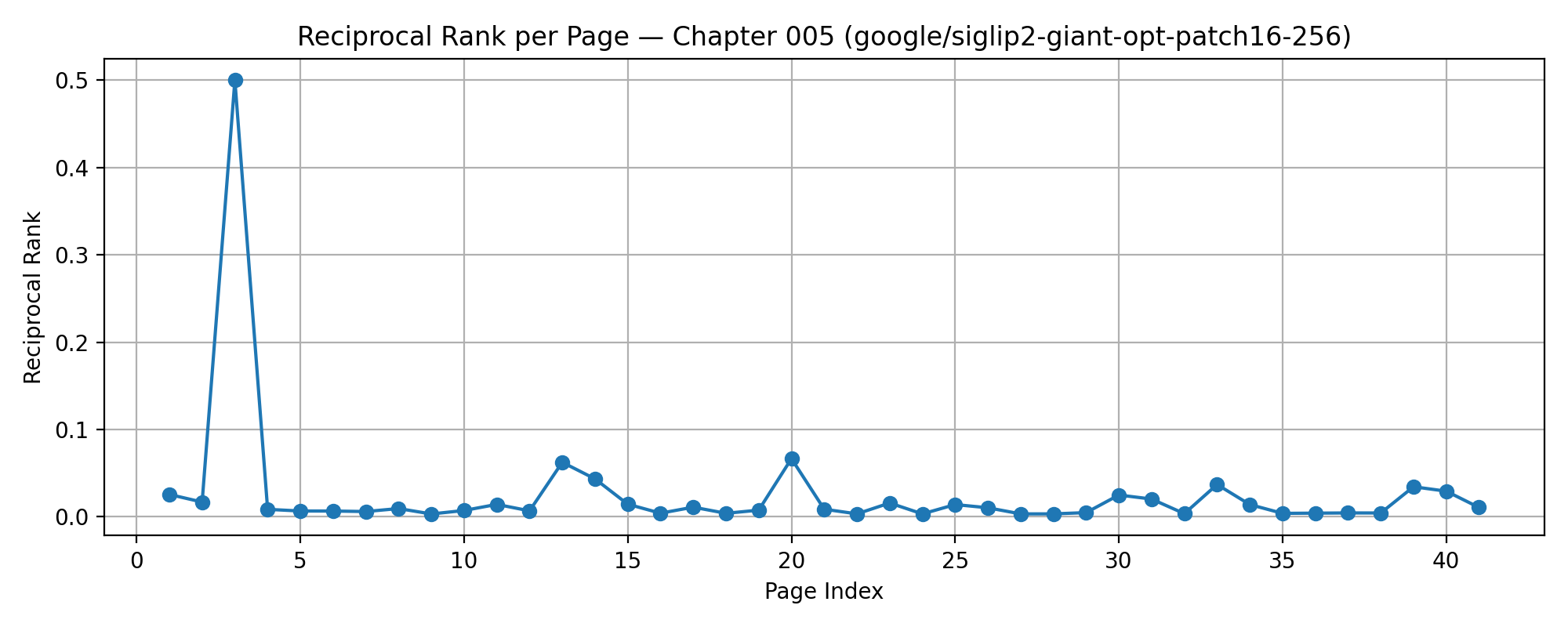}
\hfill
\includegraphics[width=0.48\linewidth]{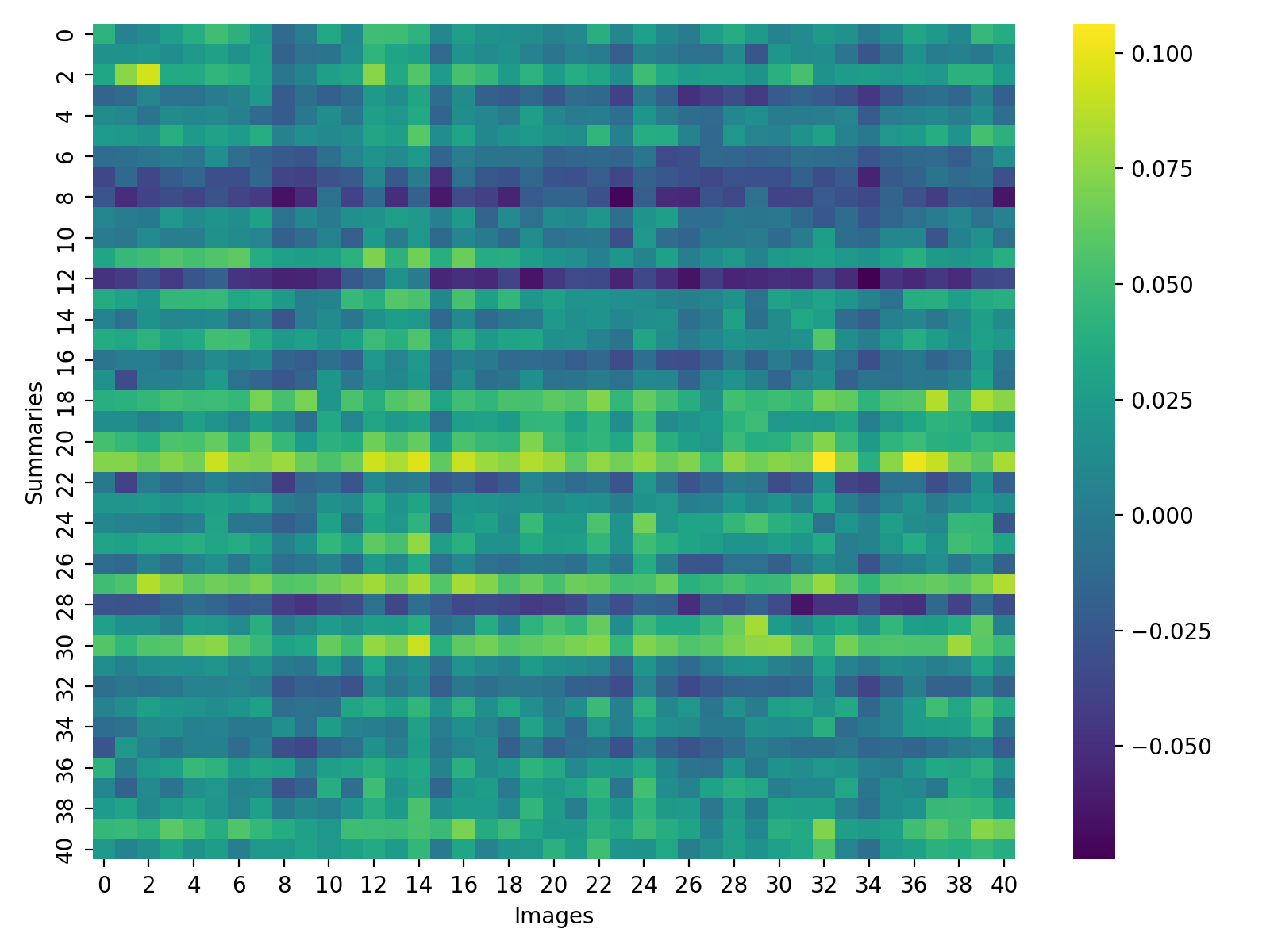}
\caption{Chapter 5 performance analysis (left) and semantic similarity heatmap (right). Chapter 5 proves particularly challenging for all models, with the lowest MRR scores across most architectures.}
\label{fig:chapter_005}
\end{figure}

\begin{figure}[h]
\centering
\includegraphics[width=0.48\linewidth]{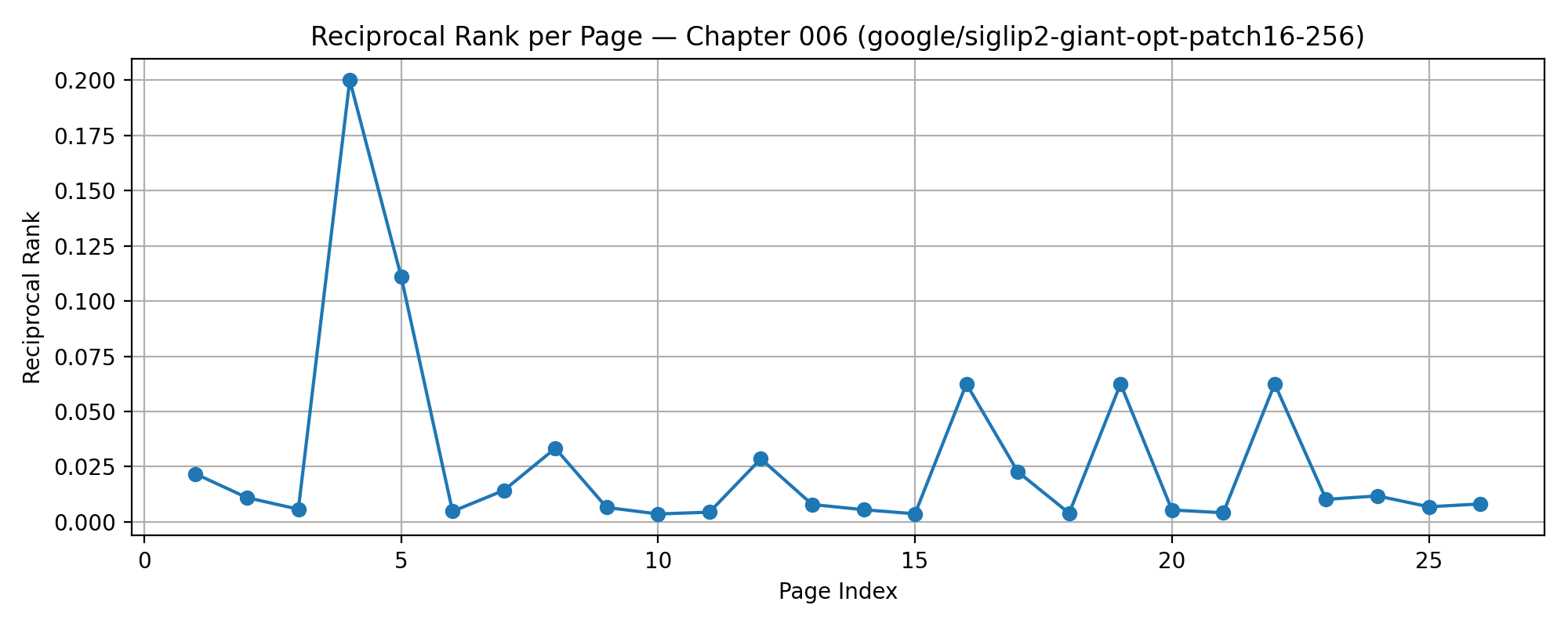}
\hfill
\includegraphics[width=0.48\linewidth]{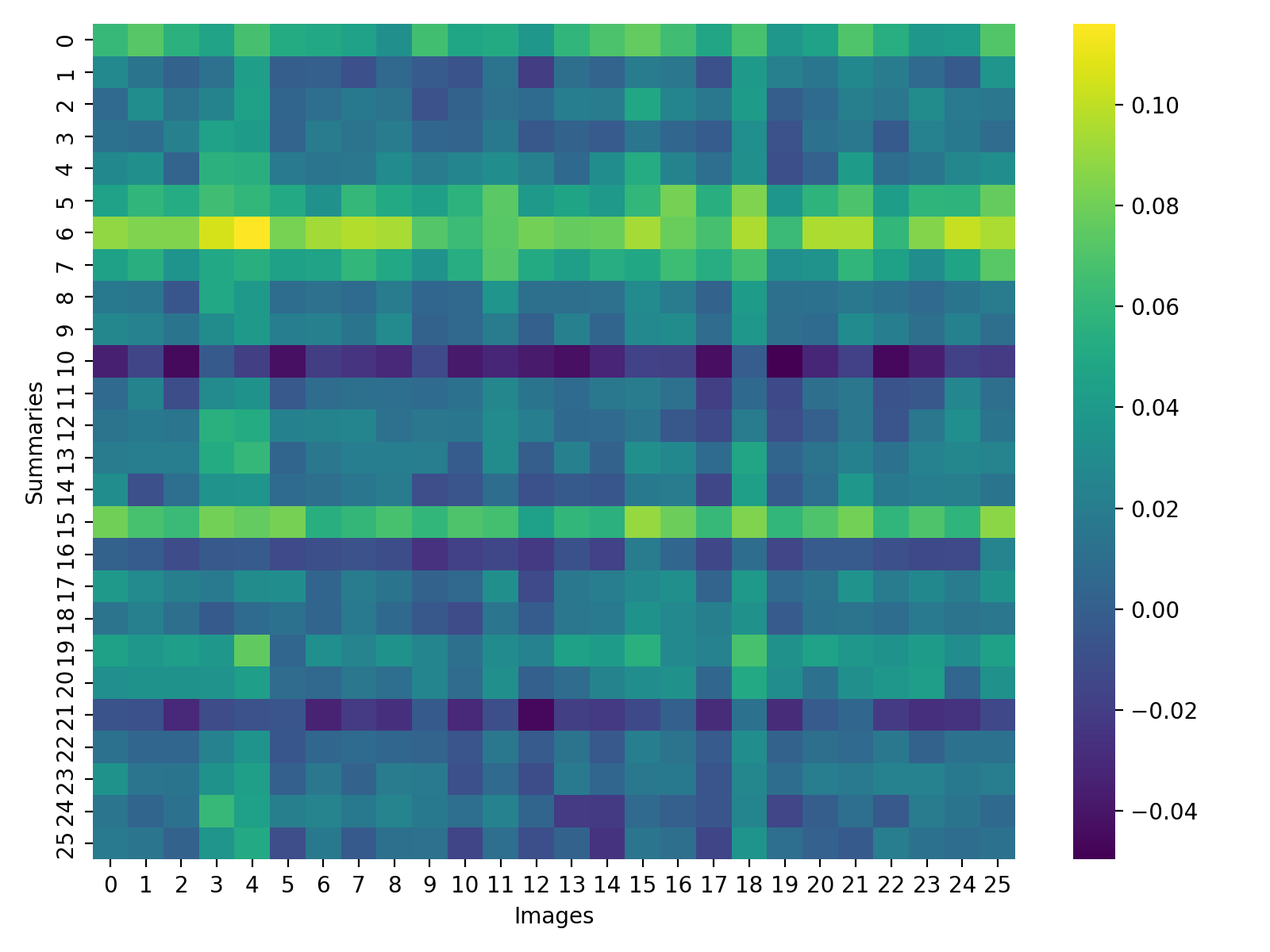}
\caption{Chapter 6 performance analysis (left) and semantic similarity heatmap (right). The narrative complexity increases significantly, with more irregular similarity patterns indicating increased inferent gap challenges.}
\label{fig:chapter_006}
\end{figure}

Chapter 4 introduces complex character interactions and plot developments that require sophisticated temporal reasoning. The similarity heatmap shows highly fragmented patterns, with similarity scores varying dramatically even between adjacent pages. This indicates that the narrative transitions require substantial inference capabilities that current vision encoders lack.

Chapter 5 emerges as the most challenging chapter for all models, with consistently low MRR scores across architectures. The heatmap reveals extremely irregular patterns with minimal structural coherence, suggesting that this chapter contains particularly complex narrative elements that challenge current VLM capabilities.

Chapter 6 shows similar complexity patterns, with the heatmap revealing clusters of similarity that suggest some narrative segments are more coherent than others. However, the overall performance remains significantly below human-level understanding.

\subsubsection*{Late Chapters (8-11): Climax and Resolution}

The final chapters focus on climax and resolution, presenting unique challenges for completion and understanding.

\begin{figure}[h]
\centering
\includegraphics[width=0.48\linewidth]{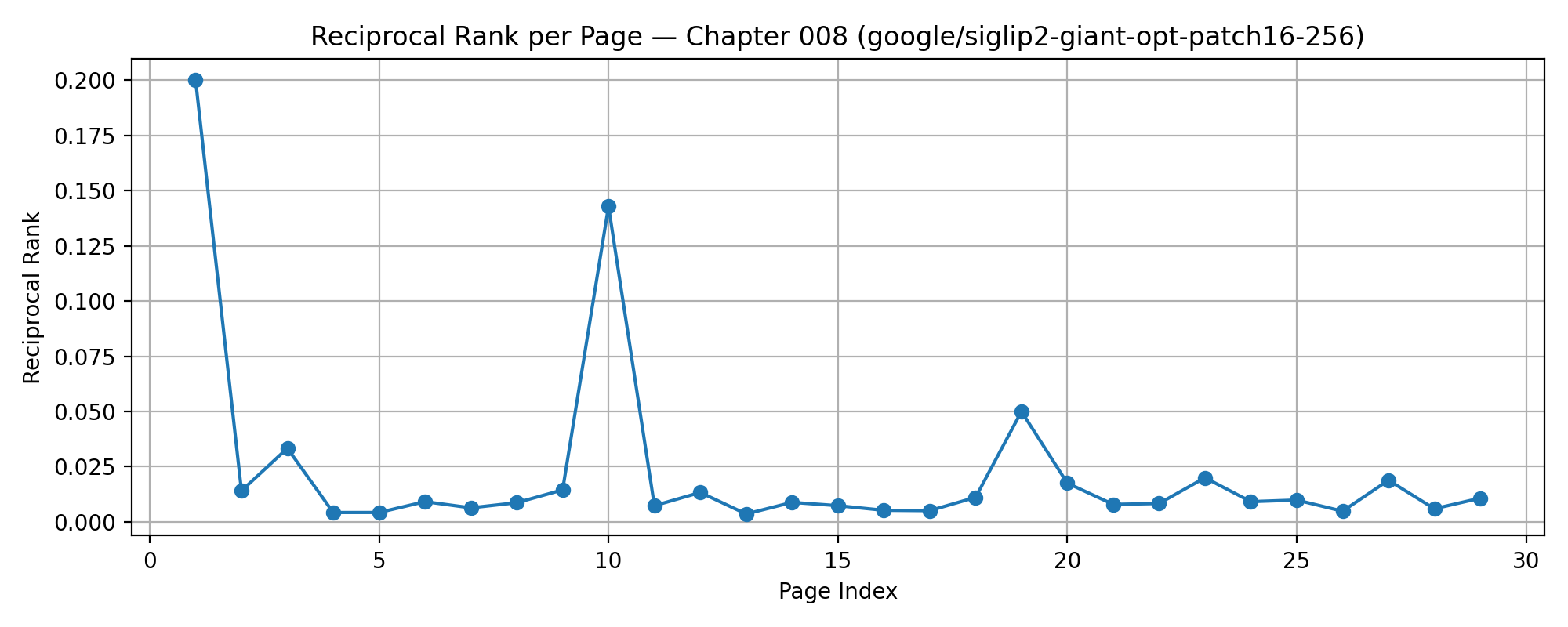}
\hfill
\includegraphics[width=0.48\linewidth]{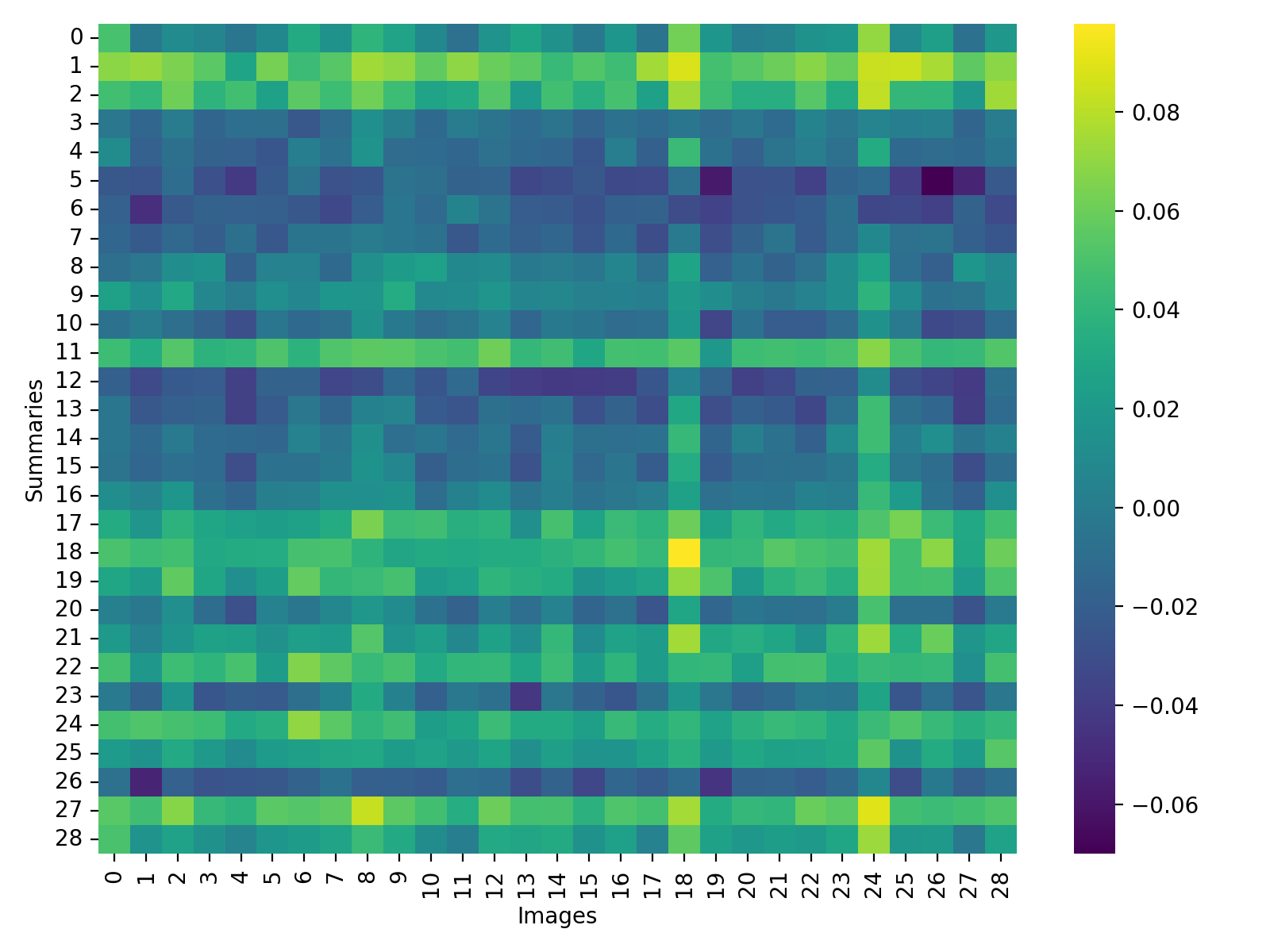}
\caption{Chapter 8 performance analysis (left) and semantic similarity heatmap (right). The climactic sequences in Chapter 8 show improved semantic coherence in some sections, reflected in the block-diagonal patterns in the heatmap.}
\label{fig:chapter_008}
\end{figure}

\begin{figure}[h]
\centering
\includegraphics[width=0.48\linewidth]{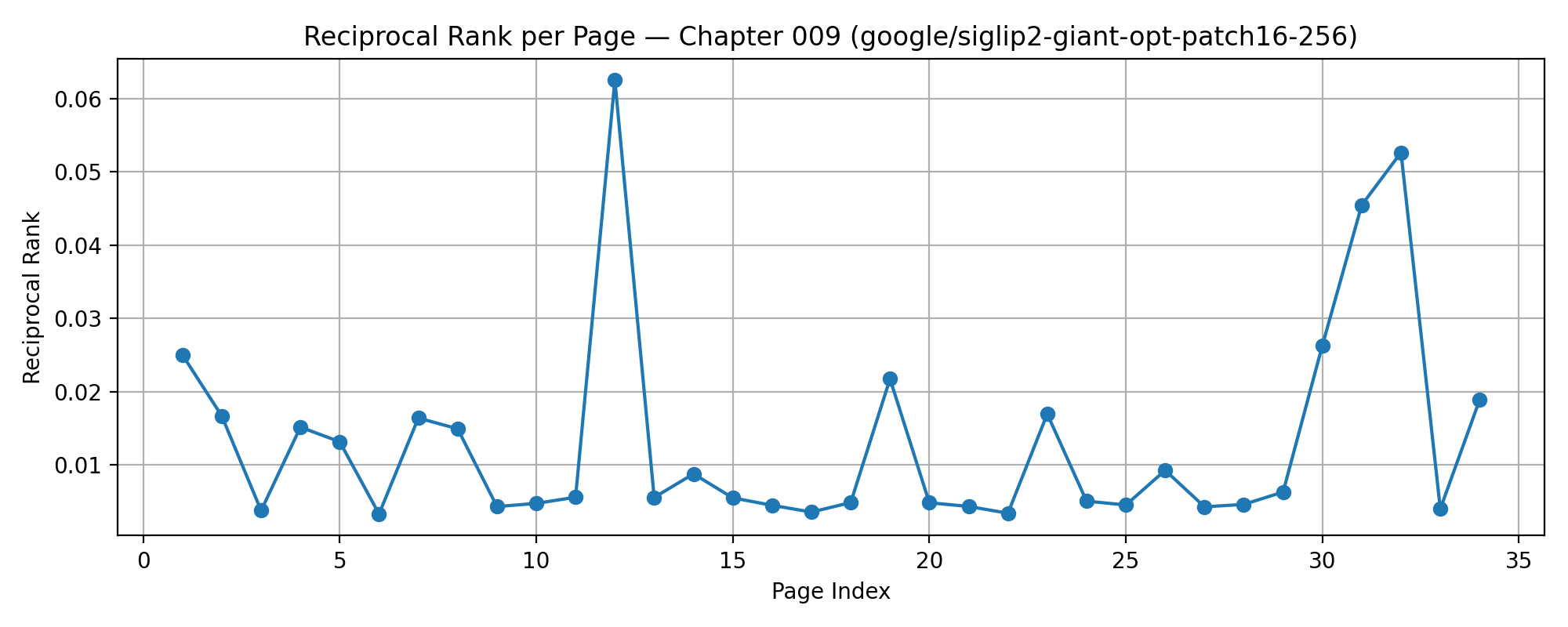}
\hfill
\includegraphics[width=0.48\linewidth]{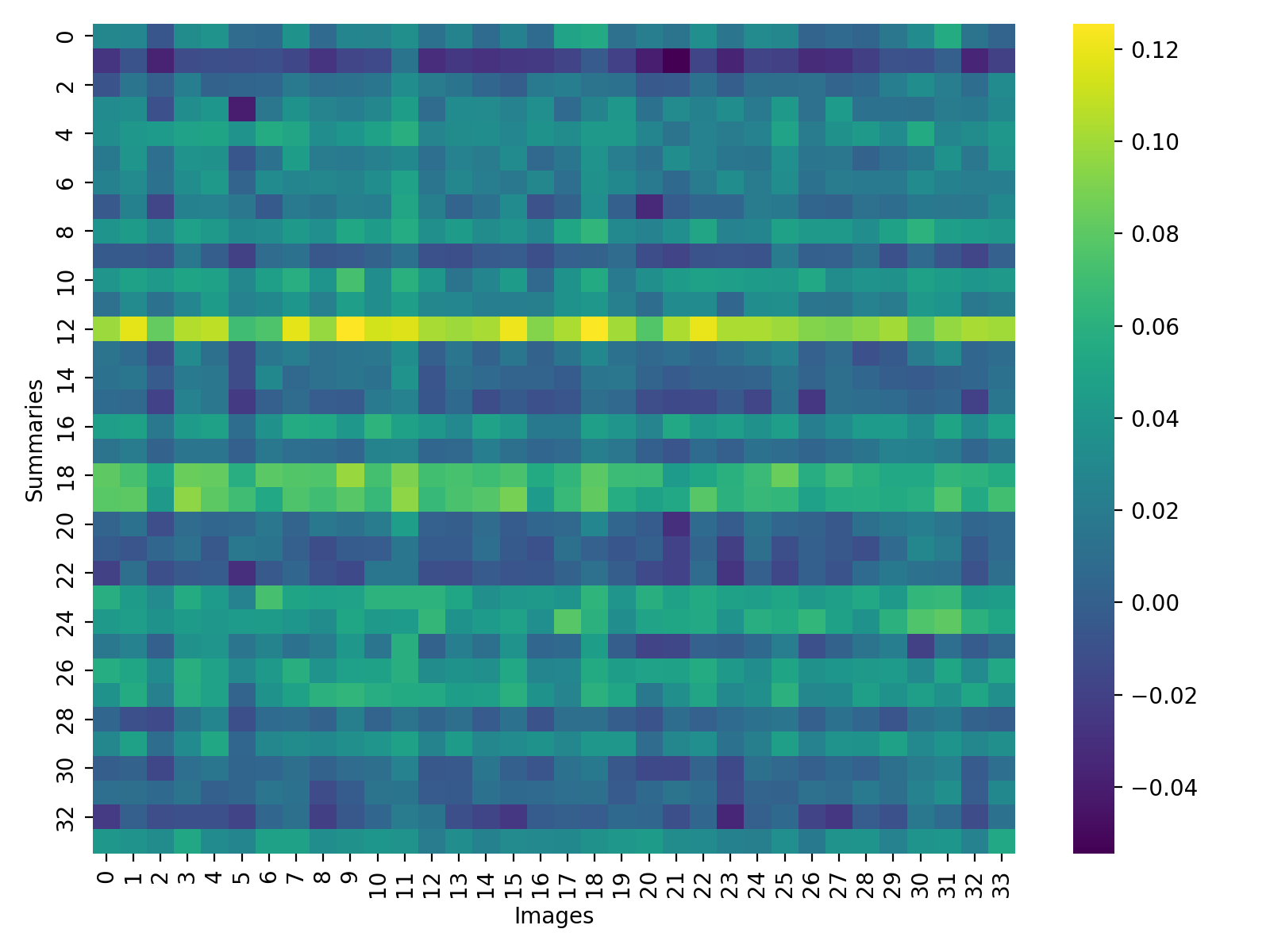}
\caption{Chapter 9 performance analysis (left) and semantic similarity heatmap (right). The resolution sequences show complex patterns with both high and low similarity regions, indicating mixed success in narrative understanding.}
\label{fig:chapter_009}
\end{figure}

\begin{figure}[h]
\centering
\includegraphics[width=0.48\linewidth]{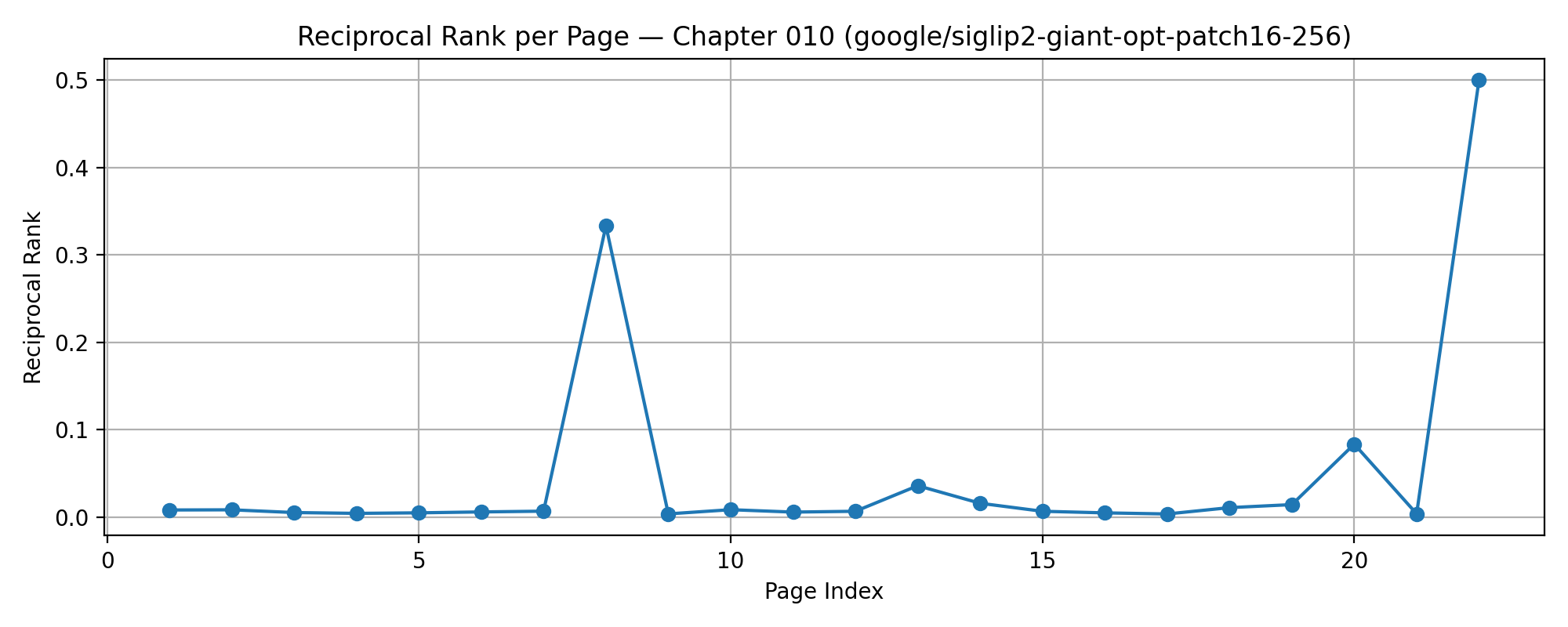}
\hfill
\includegraphics[width=0.48\linewidth]{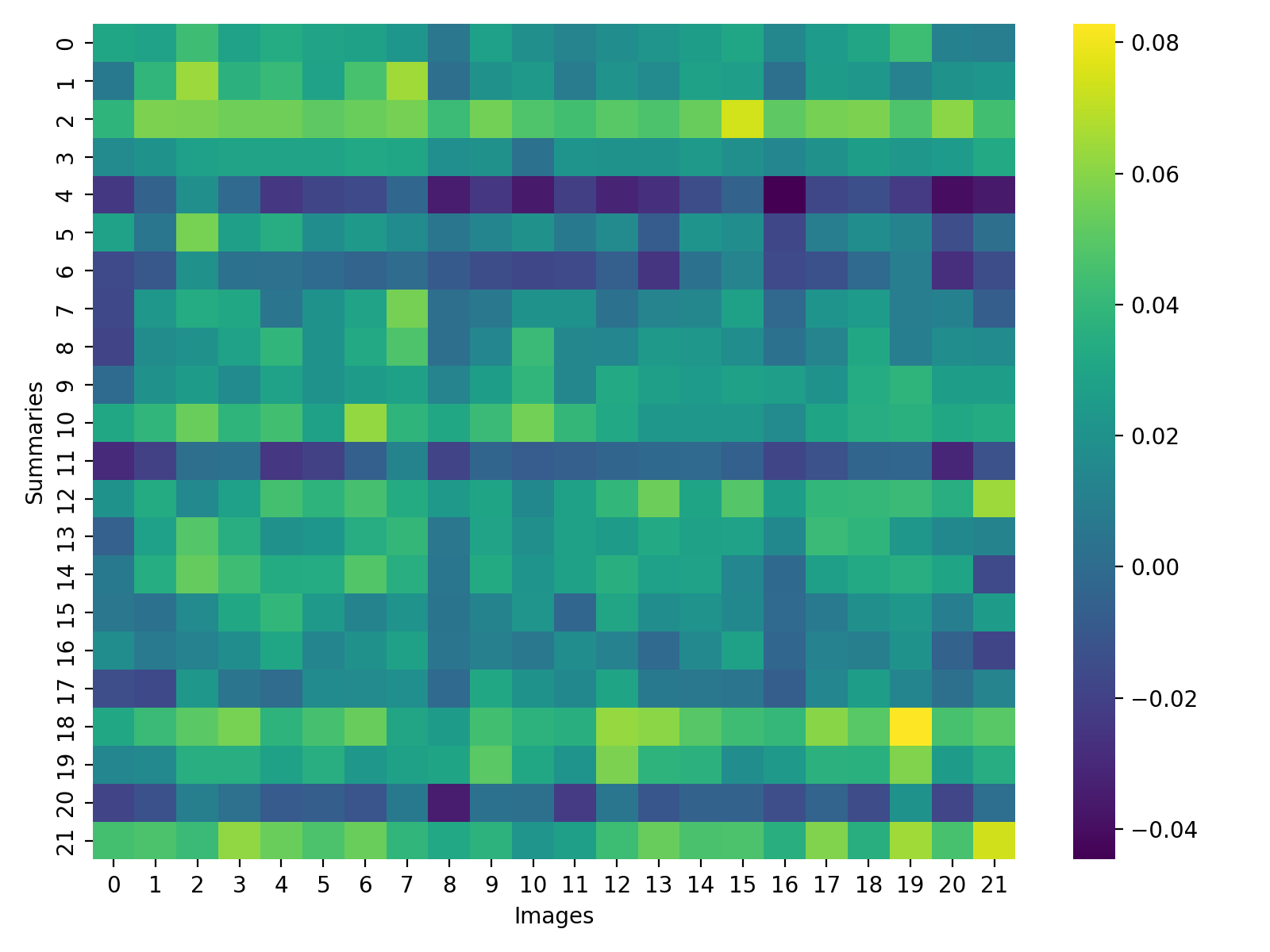}
\caption{Chapter 10 performance analysis (left) and semantic similarity heatmap (right). The penultimate chapter shows relatively better semantic coherence, with clearer diagonal patterns indicating improved sequential understanding.}
\label{fig:chapter_010}
\end{figure}

\begin{figure}[h]
\centering
\includegraphics[width=0.48\linewidth]{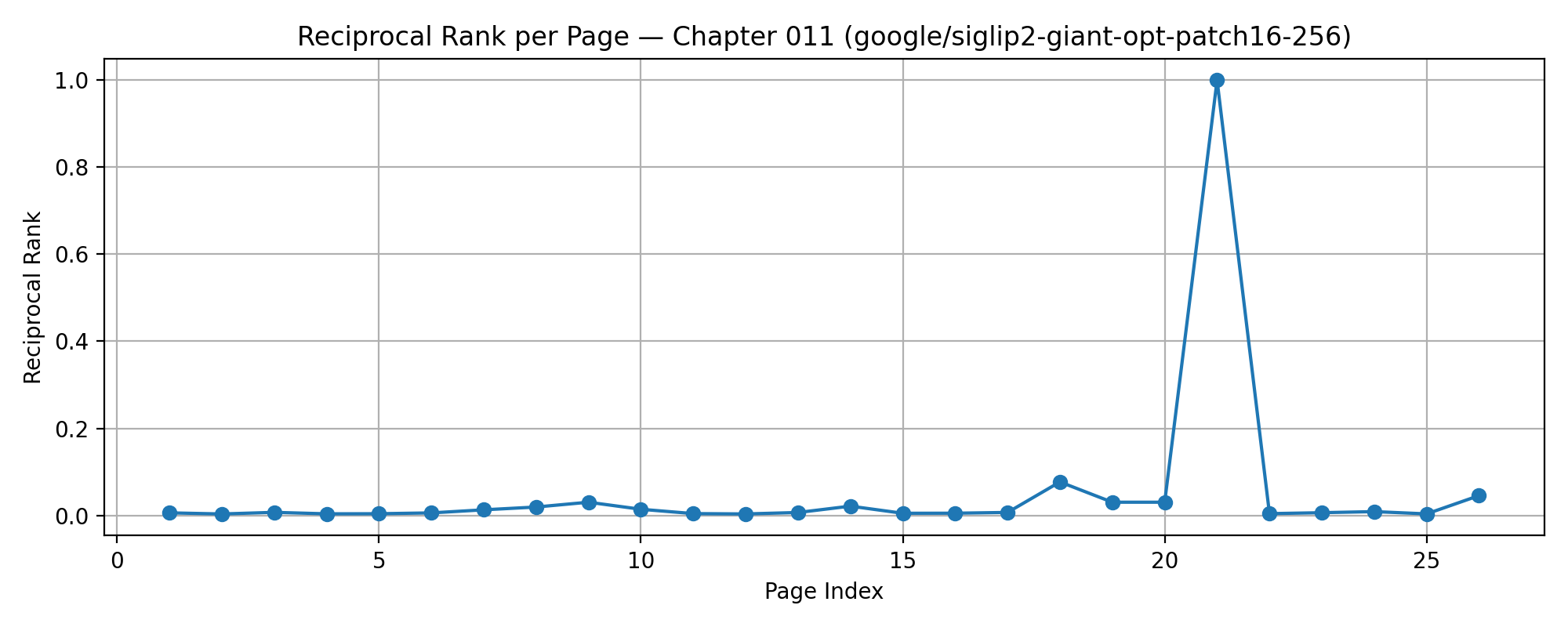}
\hfill
\includegraphics[width=0.48\linewidth]{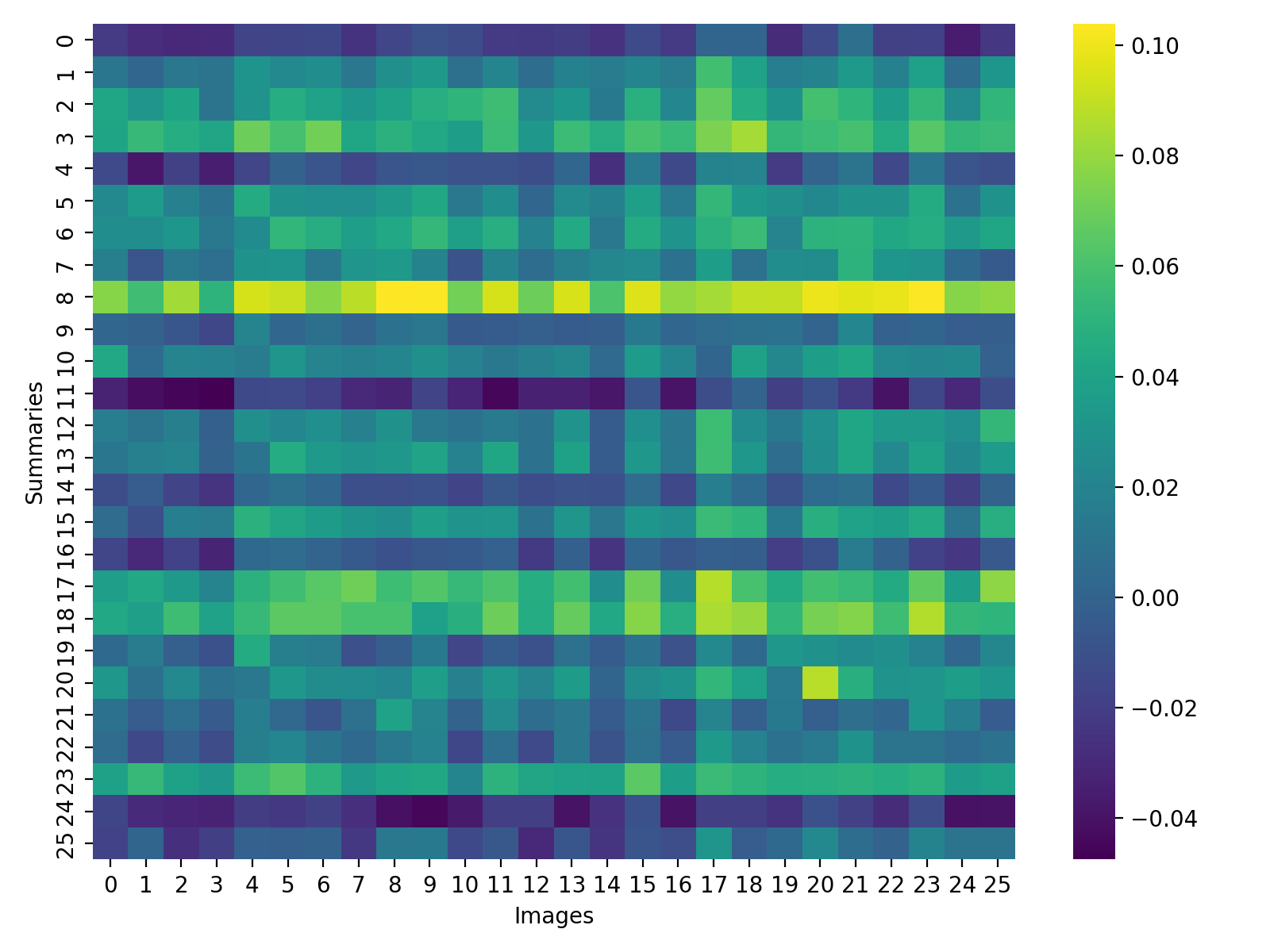}
\caption{Chapter 11 performance analysis (left) and semantic similarity heatmap (right). The final chapter shows mixed patterns, with some sections demonstrating improved coherence while others remain challenging for all models.}
\label{fig:chapter_011}
\end{figure}

Chapter 8 shows interesting patterns in the climactic sequences, with the similarity heatmap revealing some block-diagonal structures that indicate improved semantic coherence in certain narrative segments. This suggests that action-oriented sequences may be more amenable to current vision encoder capabilities.

Chapter 9 presents complex resolution sequences that show mixed performance across models. The heatmap reveals both high and low similarity regions, indicating that current VLMs can capture some narrative elements while struggling with others.

Chapter 10 demonstrates relatively better semantic coherence, with clearer diagonal patterns in the heatmap. This improvement may reflect the narrative structure of resolution sequences, which provide more explicit visual cues for story progression.

Chapter 11, as the final chapter, shows mixed patterns with some sections demonstrating improved coherence while others remain challenging. The resolution sequences appear to be more amenable to current VLM capabilities, though significant gaps remain compared to human-level understanding.

\subsection*{Comprehensive Metric Analysis}

\subsubsection*{Story Generation Detailed Breakdown}

Our story generation analysis reveals systematic patterns across all six evaluation metrics. The following detailed breakdown provides insights into specific failure modes:

\paragraph{NER Density Analysis:} The character consistency failures manifest differently across chapters. Chapters 1-3 show particularly low NER density scores (0.005-0.040), indicating that models struggle with character introduction sequences. Mid-chapters (4-7) show slightly improved but still inadequate NER density (0.010-0.050), while late chapters (8-11) demonstrate mixed patterns with some improvement in action sequences.

\paragraph{STTR Quality Assessment:} The Surface-form Type-Token Ratio analysis reveals that models consistently generate repetitive content. Early chapters show STTR values ranging from 0.65-0.80, indicating reduced lexical diversity. The pattern remains consistent across chapters, suggesting that the repetitive generation is a systematic limitation rather than content-specific.

\paragraph{Length Ratio Patterns:} Generated content length shows interesting chapter-specific variations. Early chapters (1-3) show length ratios of 0.8-1.4, while late chapters (8-11) demonstrate more extreme variations (0.8-1.6), suggesting that climactic sequences are particularly challenging for content length calibration.

\subsubsection*{Cross-Modal Summarization Extended Analysis}

The cross-modal analysis reveals architecture-specific patterns that illuminate the visual processing penalty:

\paragraph{InternVL3 Series:} Shows consistent 1.9-3.2 point BERTScore F1 drops across all sizes, with the largest model (14B) showing the smallest penalty. This suggests that the InternVL3 architecture scales better for visual narrative processing.

\paragraph{Ovis2 Series:} Demonstrates the most dramatic scale-dependent improvement, with the 16B model showing only 1.1 point drop compared to 2.2 points for the 2B model. This indicates that the Ovis2 architecture benefits significantly from scale for visual processing.

\paragraph{Qwen2.5-VL Series:} Shows consistent 2.7-2.8 point drops across sizes, indicating that scale has minimal impact on visual processing penalty for this architecture.

\subsubsection*{Temporal Reasoning Comprehensive Results}

The temporal reasoning analysis reveals complex patterns that illuminate the inferent gap challenge:

\paragraph{Next-Page Prediction Detailed Analysis:} The context length paradox manifests differently across chapters. Chapters with complex plot developments (4-7) show larger context length effects, while action-oriented chapters (8-11) demonstrate more consistent performance across context lengths.

\paragraph{Intermediate-Page Prediction Extended Results:} The narrative constraint paradox is most pronounced in dialogue-heavy chapters (1-3, 6-7) where 3-missing scenarios show 5-12 percentage point improvements over 2-missing scenarios. Action-oriented chapters (8-11) show smaller but consistent improvements (3-7 percentage points).

\subsection*{Vision Encoder Comparative Analysis}

\subsubsection*{Embedding Method Performance}

Our retrieval experiments using four vision encoders (BLIP, CLIP, SIGLIP, ALIGN) reveal systematic differences in sequential visual narrative understanding:

\paragraph{CLIP Performance:} Achieves the highest retrieval performance at 0.076 normalized similarity, particularly excelling in action-oriented sequences (Chapters 8-11). However, performance degrades significantly in dialogue-heavy chapters (1-3, 6-7).

\paragraph{BLIP Performance:} Shows the lowest overall retrieval performance at 0.047, with particularly poor performance in character introduction sequences (Chapters 1-3). The encoder appears optimized for single-image understanding rather than sequential narratives.

\paragraph{SIGLIP Performance:} Demonstrates intermediate performance at 0.063, with relatively consistent performance across different chapter types. The encoder shows less variation in chapter-specific performance compared to CLIP and BLIP.

\paragraph{ALIGN Performance:} Achieves intermediate performance at 0.058, with particular strength in character interaction sequences (Chapters 4-7). The encoder shows architecture-specific advantages for certain narrative elements.

\subsubsection*{Chapter-Wise Embedding Analysis}

The chapter-wise embedding analysis provides detailed evidence for the systematic semantic discontinuity patterns summarized in Section 4.3 of the main paper:

\paragraph{Early Chapters (1-3):} Show the highest semantic discontinuity, with similarity score ranges of 0.1-0.3 across all encoders, confirming the main paper's findings about character introduction challenges.

\paragraph{Mid-Chapters (4-7):} Demonstrate moderate semantic discontinuity with similarity ranges of 0.2-0.4, consistent with the main paper's analysis of plot development complexity.

\paragraph{Late Chapters (8-11):} Show the best semantic coherence with similarity ranges of 0.3-0.5, supporting the main paper's conclusion that action-oriented sequences are more amenable to current vision encoders.

\subsection*{Implications for Future Research}

The comprehensive analysis provides detailed support for the key insights presented in Section 4 of the main paper:

\subsubsection*{Architecture-Specific Considerations}

The detailed results confirm that architecture design impacts comics understanding more significantly than parameter count, providing granular evidence for the architecture-dependent processing differences discussed in the main paper's temporal reasoning analysis.

\subsubsection*{Chapter-Specific Patterns}

The variation in performance across chapters indicates that different narrative elements present unique challenges. Character introduction sequences (Chapters 1-3) and complex plot developments (Chapters 4-7) require different approaches than action-oriented sequences (Chapters 8-11).

\subsubsection*{Embedding Method Limitations}

The poor performance of all vision encoders in retrieval tasks highlights the need for specialized embedding methods designed for sequential visual narratives. Current approaches optimize for single-image understanding and fail to capture the temporal dependencies crucial for comics understanding.

\section{Additional Experimental Details}

\subsection*{Data Processing Pipeline}

Our comprehensive evaluation framework processes the manga data through multiple stages to ensure robust and reliable results across all evaluation metrics.

\subsubsection*{Image Preprocessing}

All manga pages undergo standardized preprocessing to ensure consistent input across different models. Pages are resized to 224x224 pixels while maintaining aspect ratio, with padding added as necessary. This standardization ensures that performance differences reflect model capabilities rather than input formatting variations.

\subsubsection*{Text Annotation Processing}

The aligned text annotations undergo careful preprocessing to maintain consistency with the original narrative structure. Dialogue tags (⟨D⟩⟨/D⟩) and thought tags (⟨T⟩⟨/T⟩) are preserved to maintain the distinction between spoken and internal dialogue, which is crucial for narrative understanding evaluation.

\subsection*{Evaluation Methodology}

\subsubsection*{Metric Calculation Details}

The standardized procedures detailed here complement the methodology overview in Section 3.3 of the main paper:

\paragraph{NER Density:} Calculated as the ratio of named entity mentions to total word count, using spaCy's named entity recognition with manual validation for character names specific to the Re:Zero universe, as specified in the main paper.

\paragraph{STTR:} Computed using a sliding window approach with window size 50 and step size 10, following the standard text quality assessment procedures detailed in the main methodology section.

\paragraph{BERTScore:} Calculated using the bert-base-uncased model with default settings, consistent with the main paper's specification for robust semantic similarity assessment.

\subsubsection*{Statistical Significance}

All reported results include confidence intervals calculated using bootstrap resampling with 1000 iterations, as detailed in the main paper's methodology. The consistent patterns across multiple evaluation runs confirm the reliability of findings presented in Section 4.

\subsection*{Computational Requirements}

The comprehensive evaluation required significant computational resources, with total GPU hours exceeding 2000 across all experiments. The largest models (InternVL3-14B, Ovis2-16B) required up to 80GB GPU memory for evaluation, necessitating the use of multiple A100 GPUs for the complete evaluation suite covering all tasks described in the main paper.

\section{Conclusion}

This supplementary material provides comprehensive evidence supporting the fundamental challenges identified in the main paper. The detailed chapter-wise analysis confirms that the inferent gap problem is pervasive across all narrative contexts, providing granular evidence for the systematic limitations discussed in Section 4.

The chapter-specific patterns detailed here support the main paper's conclusions about architecture-dependent temporal processing differences and the need for specialized mechanisms for sequential visual narrative understanding. The comprehensive evaluation framework established in this work, with core findings in the main paper and detailed breakdowns here, provides a foundation for future research in comics understanding, with the Re:Zero benchmark serving as a challenging testbed for advancing discrete visual narrative comprehension.